\documentclass[runningheads]{llncs}
\usepackage{graphicx}
\usepackage{comment}
\usepackage{amsmath,amssymb} 
\usepackage{subcaption}
\usepackage{color}

\graphicspath{{image/}}
\usepackage{url}
\usepackage[hidelinks]{hyperref}
\usepackage{algorithm}
\usepackage{algorithmic}
\usepackage{booktabs}
\usepackage{multirow}
\usepackage{caption}
\usepackage{cite}
\usepackage{ulem}



\begin{document}
\pagestyle{headings}
\mainmatter
\def\ECCVSubNumber{5931}  

\title{TP-LSD: Tri-Points Based Line Segment Detector \thanks{Y. He is the corresponding author (heyijia2016@gmail.com). S. Huang and N. Ding contribution was made when they were interns at Megvii Research Beijing, Megvii Technology, China.}}

\titlerunning{TP-LSD: Tri-Points Based Line Segment Detector}
%
\author{Siyu Huang\inst{1} \and
Fangbo Qin\inst{2} \and
Pengfei Xiong\inst{1} \and \\
Ning Ding\inst{1} \and
Yijia He\inst{1} \and 
Xiao Liu\inst{1}}
\authorrunning{Huang et al.}

%
\institute{\mbox{Megvii Technology \quad \and Institute of Automation, Chinese Academy of Sciences}\\
\email{\{siyuada7, dning97dn, heyijia2016\}@gmail.com \\ qinfangbo2013@ia.ac.cn \\ liuxiao@foxmail.com, xiongpengfei@megvii.com}}
\maketitle

\begin{abstract}
This paper proposes a novel deep convolutional model, Tri-Points Based Line Segment Detector (TP-LSD), to detect line segments in an image at real-time speed. The previous related methods typically use the two-step strategy, relying on either heuristic post-process or extra classifier. To realize one-step detection with a faster and more compact model, we introduce the tri-points representation, converting the line segment detection to the end-to-end prediction of a root-point and two endpoints for each line segment. TP-LSD has two branches: tri-points extraction branch and line segmentation branch. The former predicts the heat map of root-points and the two displacement maps of endpoints. The latter segments the pixels on straight lines out from background. Moreover, the line segmentation map is reused in the first branch as structural prior. We propose an additional novel evaluation metric and evaluate our method on Wireframe and YorkUrban datasets, demonstrating not only the competitive accuracy compared to the most recent methods, but also the real-time run speed up to \textbf{78 FPS} with the $320\times 320$ input.

\end{abstract}

\section{Introduction}
Compact environment description is an important issue in visual perception. For man-made environments with various flat surfaces, line segments can encode the environment structure, providing fundamental information to the upstream vision tasks, such as vanishing point estimation~\cite{vanish2002, vanish2011},  3D structure reconstruction~\cite{rescontruc2013}, distortion correction~\cite{Distortion}, and pose estimation~\cite{Pose,CPIE}.

\begin{figure}[thbp]
     \begin{center}
     \begin{subfigure}[b]{0.62\textwidth}
         \centering
         \includegraphics[width=1\linewidth]{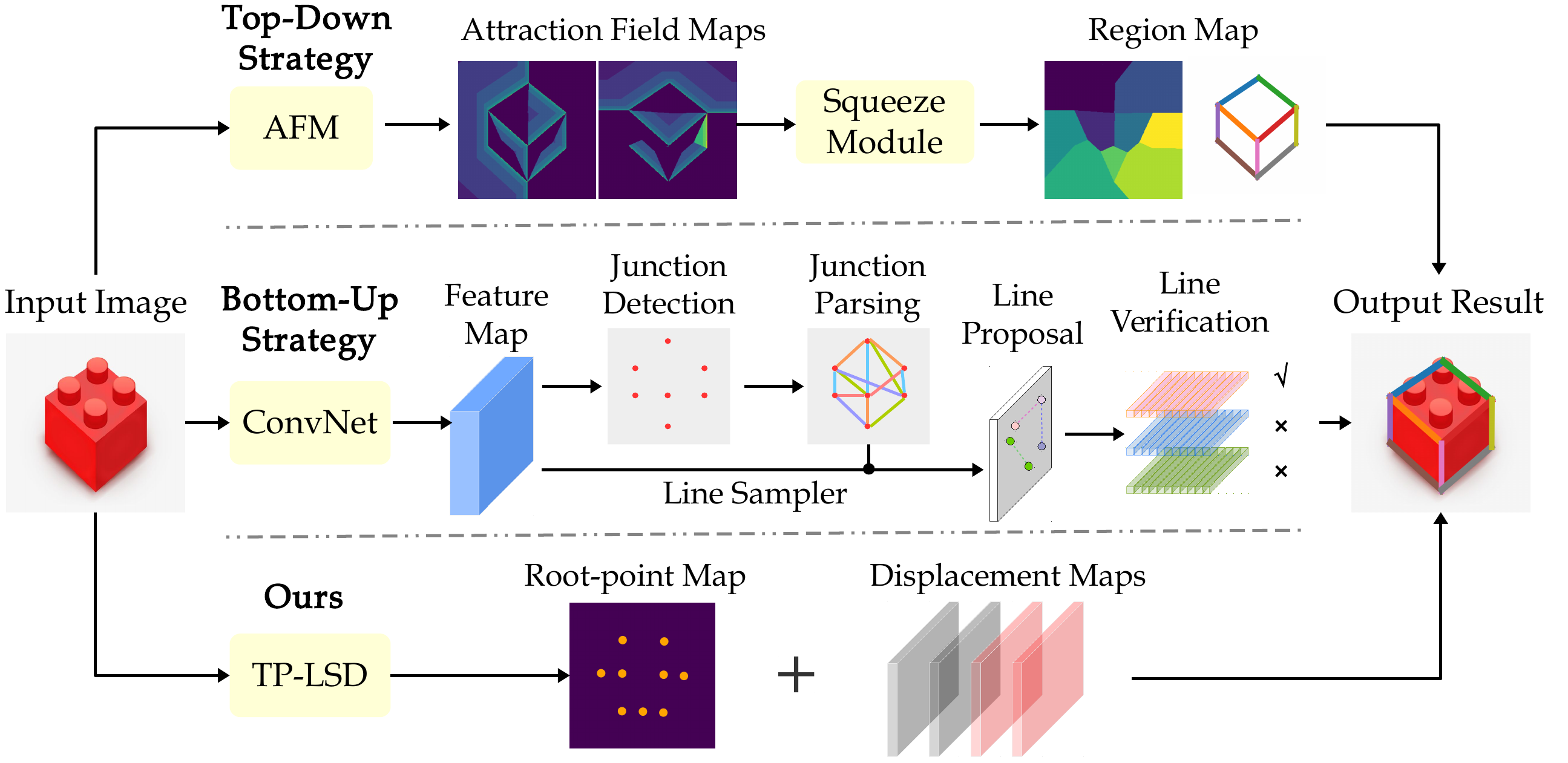}
         \caption{}
         \label{method}
     \end{subfigure}
     \begin{subfigure}[b]{0.37\textwidth}
         \centering
         \includegraphics[width=1\linewidth]{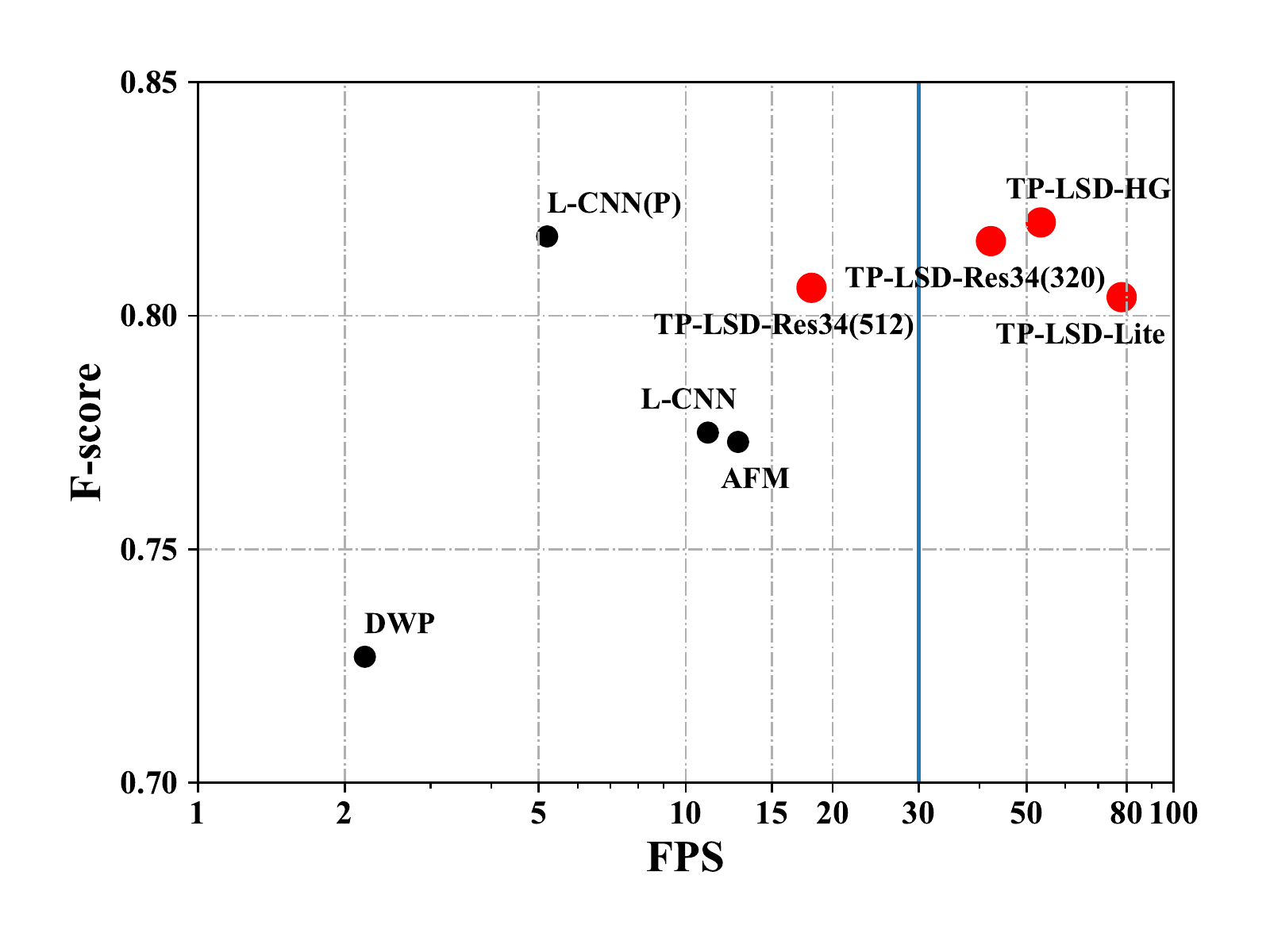}
         \caption{}
         \label{fps}
     \end{subfigure}
   \end{center}
    
    \caption{Overview. (a) Compared to the existing two-step methods, TP-LSD detects multiple line segments simultaneously in one step, providing better efficiency and compactness. (b) Inference speed and F-score on Wireframe test set.}
\end{figure}

With the rapid advance of deep learning, deep neural networks are applied to line segment detection. As shown in Fig.~\ref{method}, the existing methods have two steps. With the top-down strategy it first detects the region of a line and then squeezes the region into a line segment~\cite{AFM}, which might be affected by regional textures and does not have an explicit definition of endpoints. With the bottom-up strategy it first detect junctions and then organize them to line segments using grouping algorithm~\cite{Wireframe, Wireframe-er}, or extra classifier~\cite{PPG, LCNN, HAWP}, which might be prone to the inaccurate junction predictions caused by local ambiguity. The two-step strategy might also limit the inference speed in real-time applications.

Considering the above problems, we propose the tri-points (TP) representation, which uses a \textsl{root-point} as the unique identity to localize a line segment, and the two corresponding \textsl{end-points} are represented by their displacements w.r.t the root-point. Thus a TP encodes the length, orientation and location of a line segment. Moreover, inspired by that human perceive line segments according to straight lines, we leverage the straight line segmentation map as structural prior to guide the inference of TPs, by embedding feature aggregation modules which fuse the line-map with TP related features. Accordingly, Tri-Points Based Line Segment Detector (TP-LSD) is designed, which has three parts: feature extraction backbone, TP extraction branch, and line segmentation branch.

As to the evaluation of line segment detection, the current metrics either treat a line segment as a set of pixels, or use squared euclidean distance to judge the matching degree, which cannot reflect the various relationships between line segments such as intersection and overlapping. Therefore we propose a new metric named line matching average precision from a camera model perspective.

In summary, the main contributions of this paper are as follows:
\begin{itemize}
    \item We utilize the TP representation to encode line segment, based on which TP-LSD is proposed to realize the real-time and compact one-step detection pipeline. The synthesis of local root-point detection and global shape inference makes the detection more robust to various textures and spatial-distributions.
    \item A novel evaluation metric is designed based on the spatial imaging geometry, so that the relative spatial relationship between line segments is reflected more distinctively.
    \item Our proposed method obtains the state-of-the-art performance on two public LSD benchmarks. The average inference speed achieves up to 78 FPS, which significantly promotes the LSD applications in real-time tasks.
\end{itemize}

\section{Related Work}
\subsection{Hand-crafted Feature Based Methods}
Line segment detection is a long-standing task in computer vision. Traditional methods~\cite{EDLines, linelet, LSD, CannyLines} usually depend on low-level cues like image gradients, which are used to construct line segments with predefined rules. 
However, the hand-crafted line segment detectors are sensitive to the threshold settings and image noise. Another way to detect line segments applies Hough transform~\cite{HoughT}, which is able to use the entire image's information but difficult to identify the endpoints of line segments.

\subsection{Deep Edge and Line Segment Detection}
In the past few years, CNN-based methods have been introduced to solve the edge detection problem. HED~\cite{HED} treats edge detection problem as pixel-wise binary classification,
and achieves significant performance improvement compared to traditional methods. Following this breakthrough, numerous methods for edge detection have been proposed~\cite{RCF, BASNet}. However, edge maps lack explicit geometric information for compact environment representation. 

Most recently, CNN-based method has been realized for line segment detection.
Huang et al.~\cite{Wireframe} proposed DWP, which includes two parallel branches to predict junction map and line heatmap in an image, then merges them as line segments. Zhang et al.~\cite{PPG} and Zhou et al.~\cite{LCNN} utilize a point-pair graph representation for line segments. Their methods (PPGNet and L-CNN) first detect junctions, then use an extra classifier to create an adjacency matrix to identify whether a point-pair belongs to the same line segment. Xue e al.~\cite{AFM} creatively presented regional attraction of line segment maps, and proposed AFM to predict attraction field maps from raw images, followed by a squeeze module to produce line segments.
Furthermore, Xue et al.~\cite{HAWP} proposed a 4-D holistic attraction field map (H-AFM) to better parameterize line segments, and proposed HAWP with L-CNN pipeline.
Though learning-based methods have significant advantages over the hand-crafted ones. However, their two-step strategy might limit their real-time performance, and rely on extra classifier or heuristic post-process. Moreover, the relationship between line-map and line segments is under-utilized.

\begin{figure}[t]
     \begin{center}
     \begin{subfigure}[b]{0.24\textwidth}
         \centering
         \includegraphics[width=0.9\linewidth]{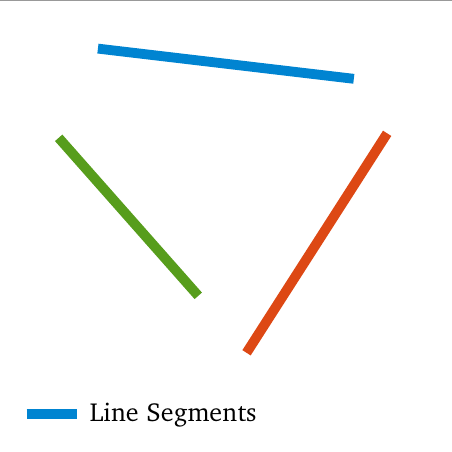}
         \caption{Pixel based}
         \label{SLS_ega}
     \end{subfigure}
     \begin{subfigure}[b]{0.24\textwidth}
         \centering
         \includegraphics[width=0.9\linewidth]{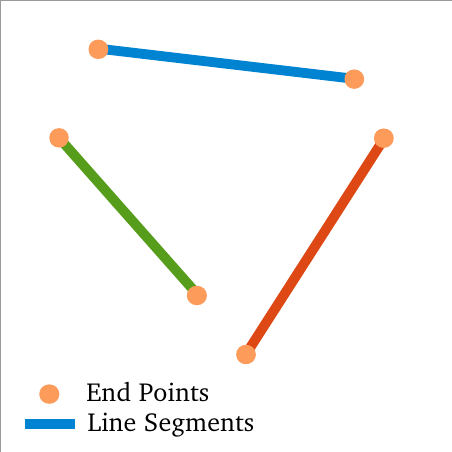}
         \caption{Endpoints based}
         \label{SLS_egb}
     \end{subfigure}
     \begin{subfigure}[b]{0.24\textwidth}
         \centering
         \includegraphics[width=0.9\linewidth]{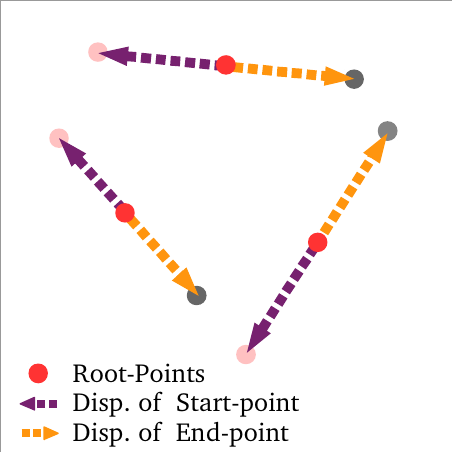}
         \caption{Tri-points based}
         \label{SLS_egc}
     \end{subfigure}
     \begin{subfigure}[b]{0.24\textwidth}
         \centering
         \includegraphics[width=0.9\linewidth]{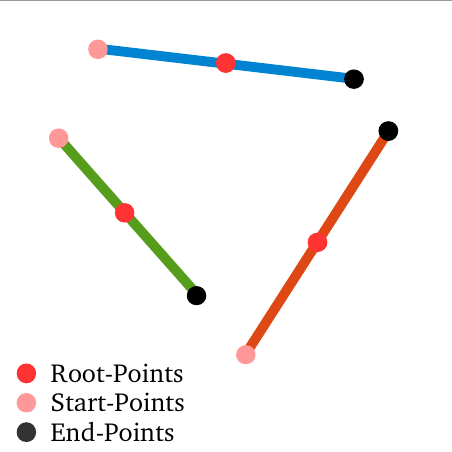}
         \caption{Vectorized lines}
         \label{SLS_egd}
     \end{subfigure}
   \end{center}
    \caption{Line segment representation.}
    
\end{figure}

\subsection{Object Detection}
 Current object detectors represent each object by an axis-aligned bounding box and classify whether its content is a specific object or background~\cite{FRcnn, SSD}. Recently, keypoint estimation has been introduced to object detection to avoid the dependence on generating boxes. CornerNet~\cite{CornerNet} detects two bounding box corners as keypoints, while ExtremeNet~\cite{ExtremeNet} detects the top-, left-, bottom-, right-most and center points of all objects. These two models both require a grouping stage to form objects based on the extracted keypoints. CenterNet~\cite{CenterNet} represents objects by the center of bounding boxes, and regresses other properties directly from image features around the center location. Such anchor-free based methods have achieved good detection accuracy with 
briefer structure, motivated by which we adopt a similar strategy to detect line segments.

\section{Tri-Points representation}
The Tri-Points (TP) representation is inspired by how people model a long narrow object. Intuitively, we usually find a root point on a line, then extend it from the root-point to two opposite directions and determine the endpoints. TP contains three key-points and their spatial relationship to encode a line segment. The root-point localizes the center of a line segment. The two end-points are represented by two displacement vectors w.r.t the root point, as illustrated in Fig.~\ref{SLS_egc},~\ref{SLS_egd}.
It is similar to SPM~\cite{SPM} used in human pose estimation.
The conversion from a TP to a vectorized line segment, which is denotes as \textbf{TP generation} opperation, is expressed by, \begin{align}
(x_{s}, y_{s}) =&(x_{r}, y_{r}) + d_s(x_r, y_r) \nonumber\\
(x_{e}, y_{e}) =&(x_{r}, y_{r}) + d_e(x_r, y_r)
\label{SLS_eq}
\end{align}
where $(x_{r}, y_{r})$ denotes the root-point of a line segment. $(x_{s}, y_{s})$ and $(x_{e}, y_{e})$ represent its start-point and end-point, respectively. Generally, the most left point is the start-point. Specially, if line segment is vertical, the upper point is the start-point. $d_s(x_r, y _r)$ and $d_e(x_r, y _r)$ denote the predicted 2D displacements from root-point to its corresponding start-point and end-point, respectively.

\begin{figure}[t]
\centering
\includegraphics[width=1\linewidth]{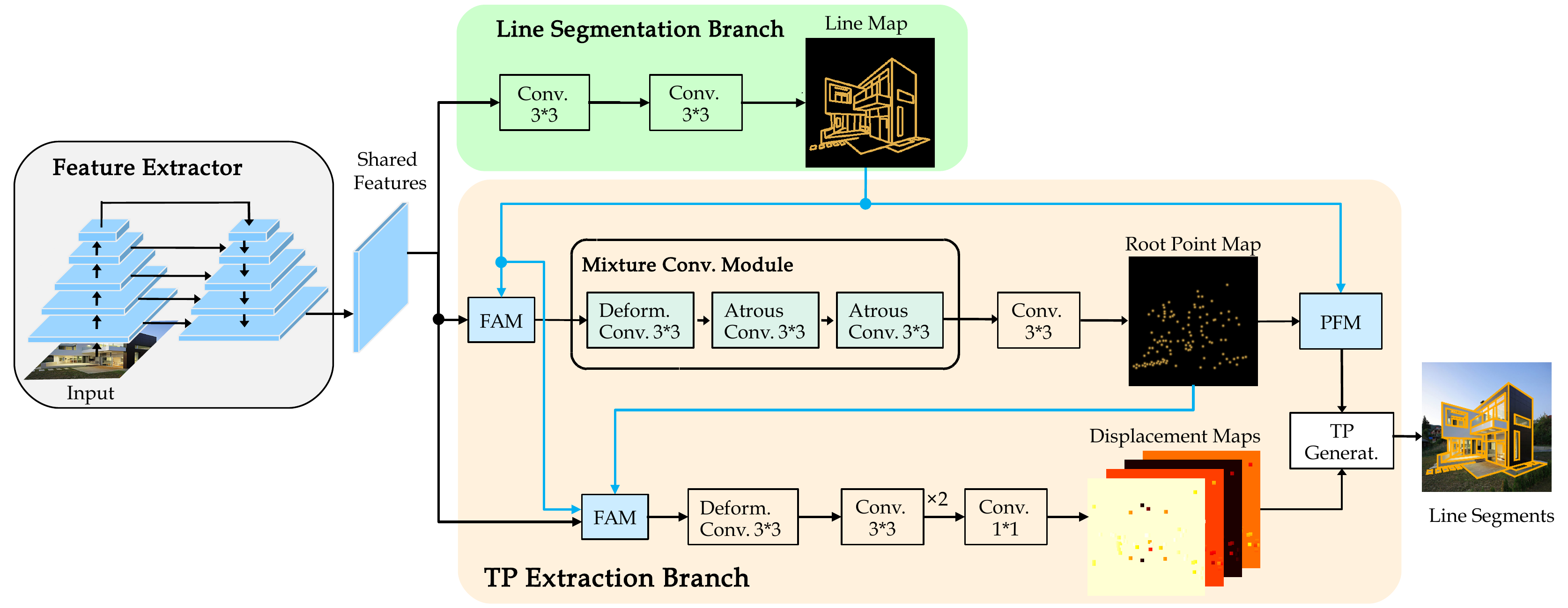}
\caption{An overview of our network architecture.}
\label{arch}
\end{figure}

\section{Methods}

Based on the proposed Tri-Points, a one-step model TP-LSD is proposed for line segment detection, whose architecture is shown in Fig.~\ref{arch}. A U-shape network is used to generate shared features, which are then fed to two branches: 1) TP extraction branch, which contains a root-point detection task and a displacement regression task; 2)line segmentation branch, which generates a pixel-wise line-map. These two branches are bridged by feature aggregation modules. Finally, after processed by point filter module, the filtered TPs are transformed to vectorized line segment instances with TP generation operation.

\subsection{TP Extraction Branch}

\textbf{Root-Point Detection}
The first task in TP extraction branch is to detect root points. Similar to CenterNet~\cite{CenterNet}, each pixel is classified to discriminate whether it is a root-point. The output activation function is sigmoid function.

\textbf{MCM.}
Because of the narrow and even long shape of line segment, it requires a large receptive field to classify the center of line segment. Therefore, a \textbf{mixture convolution module} (MCM) is introduced to provide the adaptive and expanded reception field, by cascading three convolution layers, a 3$\times$3 deformable convolutional layer, and two $3 \times 3$ atrous convolutional layers with dilation rate$=2$, whose strides are all set as 1.

\textbf{Displacement Regression}
The second task in TP extraction branch is to regress the two displacements of the start and end points w.r.t a root-point in the continuous domain. The sparse maps for the displacements are inferred by one 3$\times$3 deformable convolutional, two 3$\times$3 convolutional and a 1$\times$1 convolutional layers, whose strides are all set as 1. With the output maps, we can index the related displacements by positions. Given a root point $(x_r, y_r)$, the corresponding displacements are indexed as $d_s(x_r, y_r)$ and $d_e(x_r, y_r)$. Then the coordinates of the start- and end-points can be obtained by Eq.~\eqref{SLS_eq}.

\subsection{Line Segmentation Branch}
Pixel-wise map of straight lines is easier to obtain because the precise determination of end-points is not required. Based on the idea that line segment is highly related to straight line, we use a straight line segmentation branch to provide prior knowledge for line segment detection. First, straight line can serve as spatial attention cue. Second, a root-point must be localized on a straight line. As is shown in Fig.~\ref{arch}, the line segmentation branch has two 3$\times$3 convolutional layers with the stride 1. The output activation function is sigmoid function, so that the line-map $P(L)$ has the pixel values ranging within $(0,1)$.

\textbf{FAM.}
From the multi-modal feature fusion prospective, we present a \textbf{feature aggregating module} (FAM) to aggregate the structural prior of line-map with the TP extraction branch. Given a line-map $P(L)$ from the line segmentation branch, the straight line activation map $A_l$ is obtained by tanh($w\times P(L)+b$) where $w, b$ denotes the parameters of a $1 \times 1$ convolutional layer, and the tanh gating function indicates whether a pixel is activated or suppressed according to its relative position to a straight line. The shared feature is firstly aggregated with the straight line activation map $A_l$ by concatenation, and then fed to the root-point detection sub-branch, as shown in Fig.~\ref{arch}.
For the displacement regression sub-branch, similarly, the straight line activation map and the root-point activation map are obtained by 1$\times$1 conv and tanh, then fused with the shared feature map by concatenation, as shown in Fig.~\ref{arch}. Thus the prior knowledge of straight line and root point can benifit the displacement regression.

\textbf{PFM.}
The line-map can also be leveraged to filter the noisy root-points that lies out of line. We consider the root-point confidence map as a probability distribution $P(R|L)$ conditioned on line existence. Thus the root-point confidence map $P(R|L)$ can be refined by the multiplication with the line confidence map $P(L)$, which is called \textbf{point filter module} (PFM), as given by
\begin{equation}
\Tilde{P}(R) = \Tilde{P}(R|L) \times \Tilde{P}(L) ^ \alpha
\label{combine}
\end{equation}
where the power coefficient $\alpha\in (0,1)$ is to adjust the contribution of line-map.

\subsection{Training and Inference}\label{infer}

\textbf{Feature extractor.}
A U-shape network is used as the feature extractor. After a backbone encoder, there are four decoder blocks. Each decoder block is formed by a bi-linear interpolation based up-sampling and a residual block. Skip connection is used to aggregate multi-scale features by concatenating the low level features with the high level features. The output of the feature extractor is a 64-channel feature map, whose size is the same with the input image, or optionally half of the input size for faster inference. This feature map is used as the shared features for the following branches.

\textbf{Loss.}
In training stage, the input image is resized to $320 \times 320$, and the outputs include a line-map, a root-point confidence map, and four displacement maps, whose ground truths are generated from the raw line segment labels. The three tasks' losses are combined as Eq. \eqref{loss_all}, where $\lambda_{root, disp, line}=\{50, 1, 20\}$. \begin{equation}
\mathbb{L}_{total} = \lambda_{root}\mathbb{L}_{root} + \lambda_{disp}\mathbb{L}_{disp} + \lambda_{line}\mathbb{L}_{line}
\label{loss_all}
\end{equation}

The ground truth of root-point confidence map is constructed by marking the root-point positions on a zero-map and then smoothed by a scaled 2D Gaussian kernel truncated by a $5 \times 5$ window, so that the root-point has the highest confidence $1$, and its nearby pixels have lower confidence. A weighted binary cross-entropy loss $\mathbb{L}_{root}$ is used to supervise this task. The ground truths of the displacement maps are constructed by assigning displacement values at the root-point positions on the zero-maps. For each ground truth line segment, its mid-point is considered as the root point. For the pixels within a $5 \times 5$ window centered at the mid-point, we calculate the displacements from it to the start- and end-points, then assigned the displacement values to these pixels. After all the ground truth line segments are visited, the final displacements maps are used for smoothed L1 loss $\mathbb{L}_{disp}$ based regression learning. Note that only the root-points and its $5 \times 5$ neighbourhood window are involved in the loss calculation. As to the line segmentation sub-task, the ground truth of line segmentation map are constructed by simply draw the line segments on a zero map and the learning is supervised by the weighted binary cross entropy loss $\mathbb{L}_{line}$.

In the inference stage, after the root-point confidence map is produced, the non-maximum suppression is operated to extract the exact root-point positions.  Afterwards, we use the extracted root-points and their corresponding displacements to generate line segments from TPs with Eq. \eqref{SLS_eq}.

\section{Evaluation Metrics} \label{eva_matric}
In this section, we briefly introduce two existed evaluation metrics: pixel based metric and structural average precision, and then design a novel metric, line matching average precision.

\textbf{Pixel based metric:}
For a pixel on a detected line segment, if its minimum distance to all the ground truth pixels is within the 1 percent of the image diagonal size, it is regarded as true positive. 
After evaluating all the pixels on the detected line segments, the F-score F$^H$ can be calculated ~\cite{Wireframe, AFM, LCNN}. The limitation is that it cannot reveal the continuity of line segment. For example, if a long line segment is broken into several short ones, the F-score is high but these split line segments is not suitable for 3D reconstruction or wireframe parsing.

\begin{figure}[t]
    \centering
    \begin{subfigure}[b]{0.3\textwidth}
         \centering
         \includegraphics[width=\linewidth]{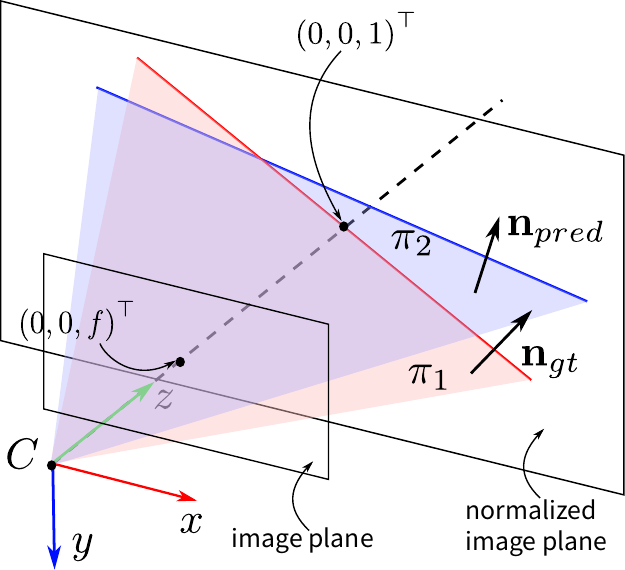}
         \caption{\centering Geometric}
         \label{sub_fig:metric}
     \end{subfigure}
     ~~
     \begin{subfigure}[b]{0.3\textwidth}
         \centering
         \includegraphics[width=\linewidth]{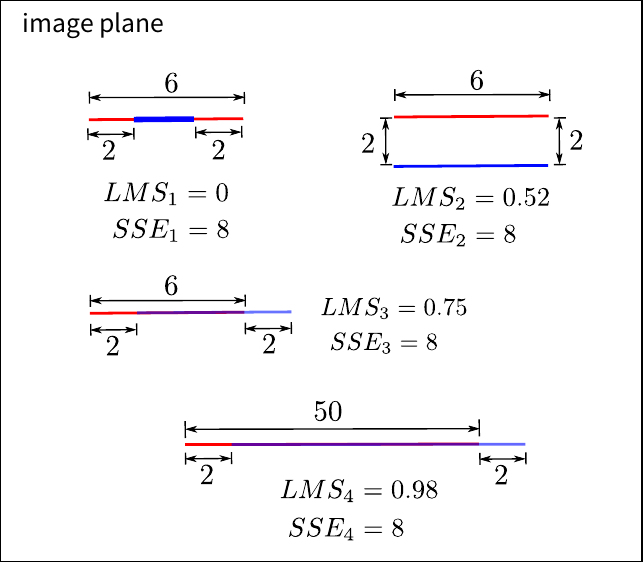}
         \caption{\centering Overlapped}
         \label{sub_fig:overlapped}
     \end{subfigure}
    ~~
    \begin{subfigure}[b]{0.3\textwidth}
         \centering
         \includegraphics[width=\linewidth]{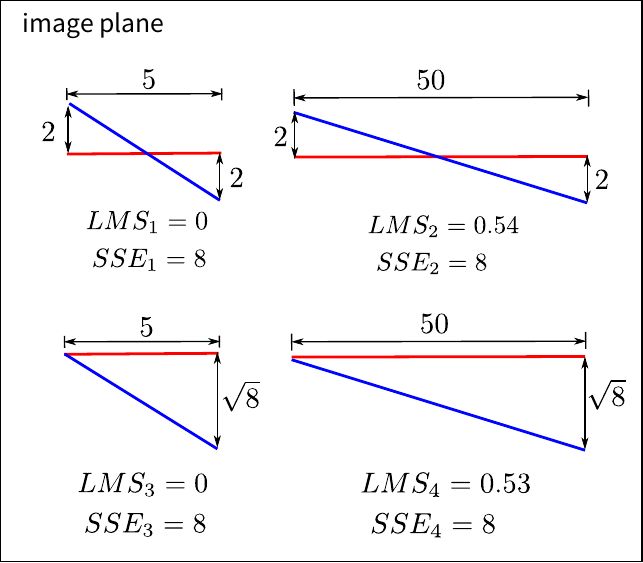}
         \caption{\centering Intersection}
         \label{sub_fig:intersection}
    \end{subfigure}

    \caption{Evaluation metrics for line segment detection. (a) The geometric explanation of the proposed line matching score (LMS). The blue and red line segments on the normalized image plane correspond to the detection and ground truth, respectively, which determine two planes together with the optical center C. 
    In (b) and (c), the different matching situations could have the same SSE score 8 with sAP metric. In contrast, the LMS gives the discriminative scores.
    }
    \label{sap}
\end{figure}

 \textbf{Structural Average Precision:} The structural average precision (sAP)~\cite{LCNN} uses 
 the sum of squared error (SSE) between the predicted end-points and their ground truths as evaluation metric. The predicted line segment will be counted as a true positive detection when its SSE is less than a threshold, such as $\epsilon=5, 10, 15$. 
 However, line segment matching could be more complicated than point pair correspondence. For example, in Fig.~\ref{sub_fig:overlapped},~\ref{sub_fig:intersection}, it is shown that sAP is not discriminative enough for some different matching situations.

\textbf{Line Matching Average Precision:}\label{line_matching_metric2}
To better reflect the various line segment matching situations in term of direction and position as well as length, the Line Matching Score (LMS) is proposed. LMS contains two parts: $Score_\theta$ denotes the differences in angle and position, and $Score_l$ denotes the matching degree in length. The LMS is calculated by 
\begin{equation}
  LMS = Score_\theta \times Score_l
\end{equation}

Inspired by 3D line reconstruction, $Score_\theta$ is calculated in the 3D camera frame as shown in Fig.~\ref{sub_fig:metric}. A line segment and the camera's optical center jointly determine a unique plane whose normal vector is $\mathbf{n}$.
Thus, given a predicted and a ground truth line segments, they determine two 3D planes, and the angle between their normal vectors is used to measure the directional matching degree. The angle is equal to 0 if and only if the two line segments are collinear. To calculate $Score_\theta$, Firstly, a ground truth line segment is aligned to the center of the image plane by subtracting the coordinates of the midpoint $\mathbf{l}_m = \left(x_m,y_m\right)^{\top}$. The endpoints of detected line segment is also subtracted by $\mathbf{l}_m$. Then, the endpoints are projected from the 2D image plane $\mathbf{l}_i=\left(x_i,y_i\right)^{\top},i=s,e$ onto the 3D normalized image plane by dividing the camera focal length, i.e. $\Bar{\mathbf{l}}_i=\left(\frac{x_i}{f},\frac{y_i}{f},1\right)^{\top}$. Finally, the normal vectors $\mathbf{n}_{gt}$ and $\mathbf{n}_{pred}$ are obtained by cross-multiplying their endpoint $\Bar{\mathbf{l}}_s \times \Bar{\mathbf{l}}_e$, respectively. $Score_\theta$ is given by,
\begin{align}
Score_\theta=
\begin{cases}
1-\frac{\theta(\mathbf{n}_{gt}, \mathbf{n}_{pred})}{\eta_{\theta}}, &\quad \text{if }\theta(\mathbf{n}_{gt}, \mathbf{n}_{pred}) < \eta_{\theta} \\
0, &\quad \text{otherwise}
\end{cases}
\end{align}
where $\theta\left(\right)$ is to calculate the angle between two vectors with the unit degree. $\eta_\theta$ is a minimum threshold.

$Score_l$ demonstrates the overlap degree of two line segment. The ratio of overlap length against the ground truth length is $\eta_1$. The ratio of overlap length against the projection length is $\eta_2$.
\begin{equation}
  \eta_1=\frac{\mathcal{L}_{pred} \cap \mathcal{L}_{gt}}{\mathcal{L}_{gt}}, ~
    \eta_2 = \frac{\mathcal{L}_{pred} \cap \mathcal{L}_{gt}  }{\mathcal{L}_{pred}\left|\cos(\alpha)\right|}
\end{equation}
where $\mathcal{L}$ is the length of line segment and $\mathcal{L}_{pred} \cap \mathcal{L}_{gt}$ is the overlap length of the predicted line segment projected to the ground truth line segment. $\alpha$ is the angle between the two line segments in 2D image. Then $Score_l$ is calculated by,
\begin{align}
Score_l=
\begin{cases}
\frac{\eta_1+\eta_2}{2}, &\quad \text{if }\eta_1 \geq \eta_{l}\text{, and } \eta_2 \geq \eta_{l} \\
0, &\quad \text{otherwise}
\end{cases}
\end{align}
where $\eta_{l}$ denotes a minimum threshold. Since the focal length of a camera might be unknown for public data sets, to make a fair comparison, we firstly re-scale the detected line segments with the same ratio of resizing the original image to the resolution $128\times128$, and set a virtual focal length $f=24$. Besides we set $\eta_\theta=10^{\circ}$ and $\eta_{l}=0.5$ in this work.

\begin{figure}[t]
     \begin{center}
     \begin{subfigure}[b]{0.23\textwidth}
         \centering
         \includegraphics[width=1\linewidth]{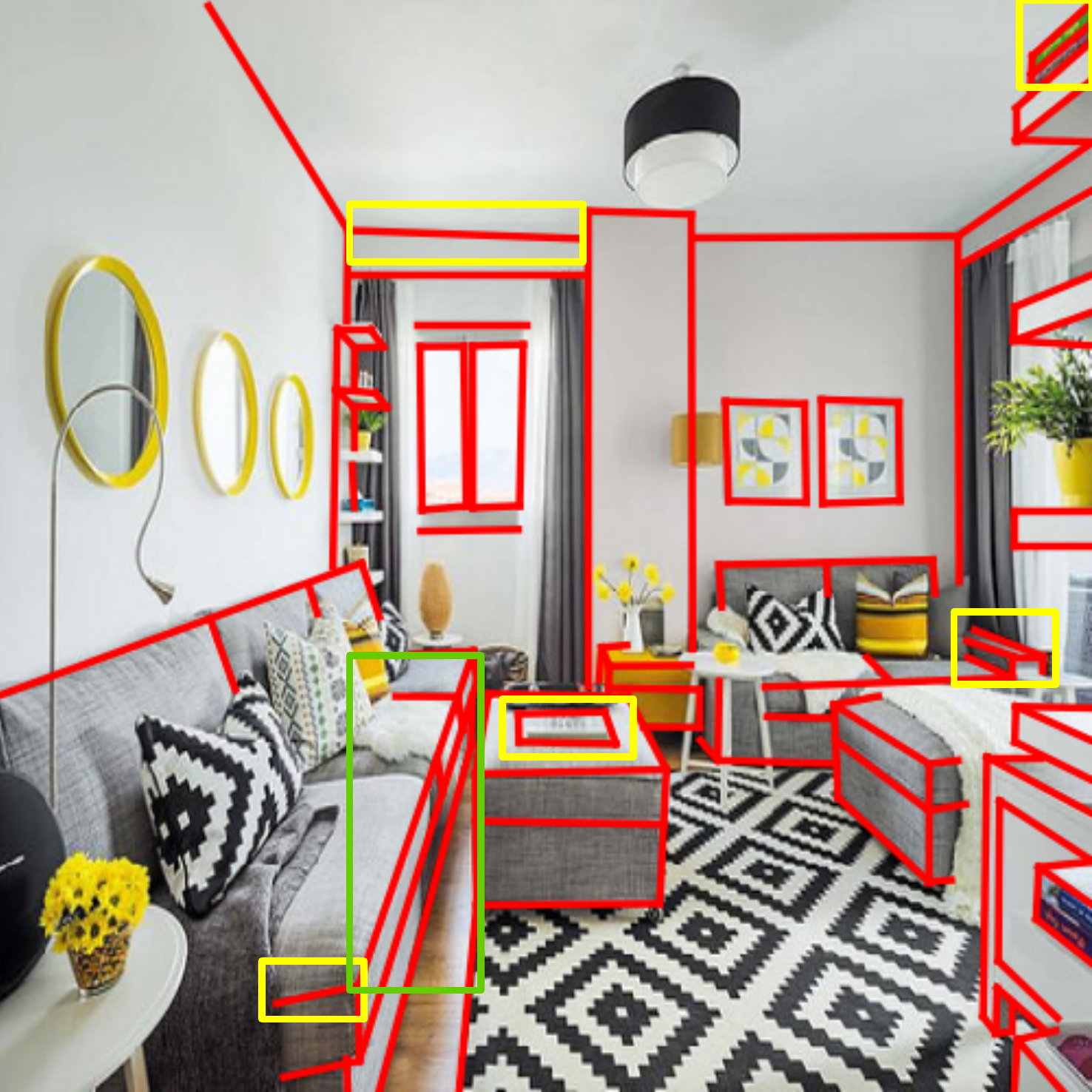}
         \caption{GT}
         \label{sub_fig:gt_match}
     \end{subfigure}
     ~~
     \begin{subfigure}[b]{0.23\textwidth}
         \centering
         \includegraphics[width=1\linewidth]{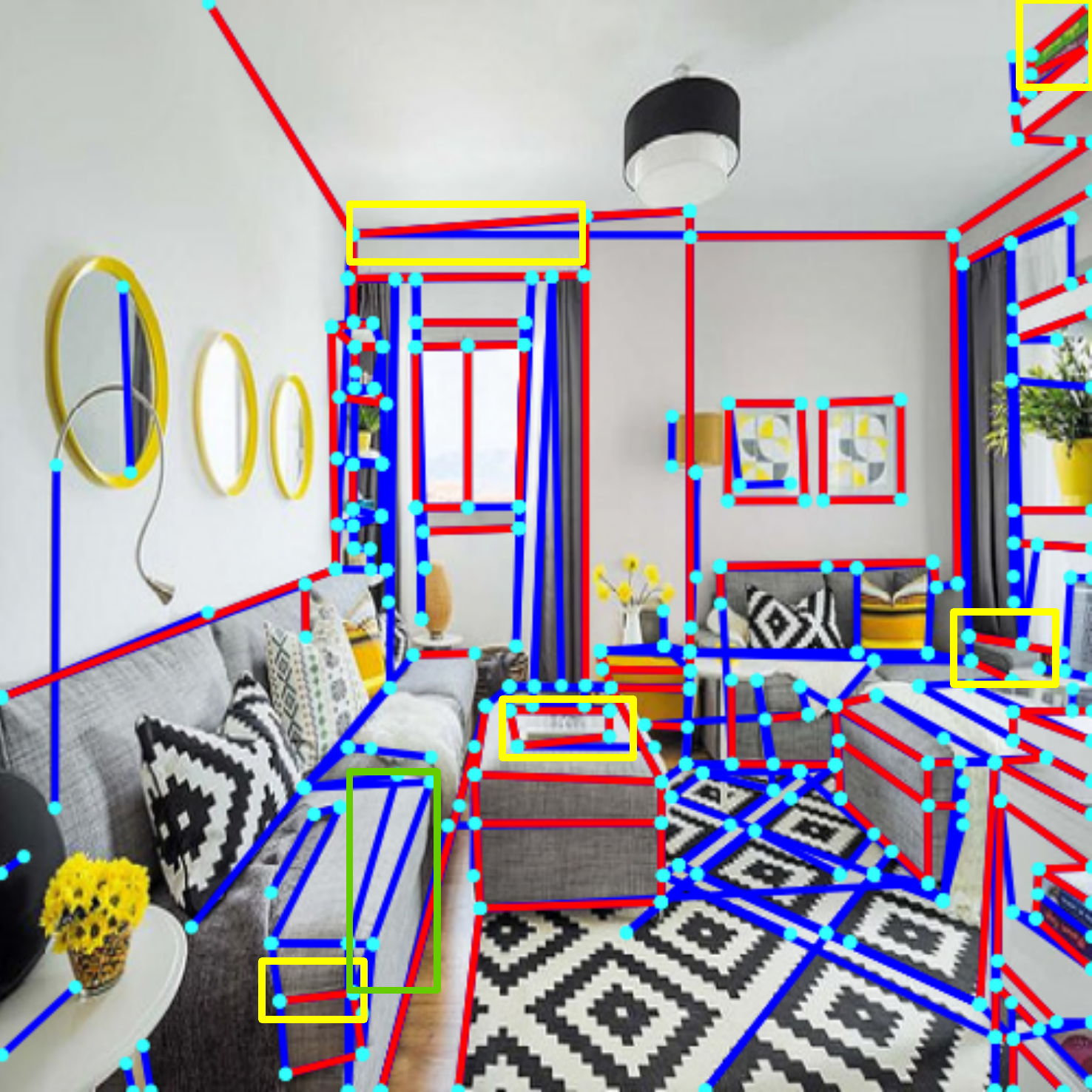}
         \caption{sAP$^{10}$ Matching}
         \label{sub_fig:sap_match}
     \end{subfigure}
    ~~
     \begin{subfigure}[b]{0.23\textwidth}
         \centering
         \includegraphics[width=1\linewidth]{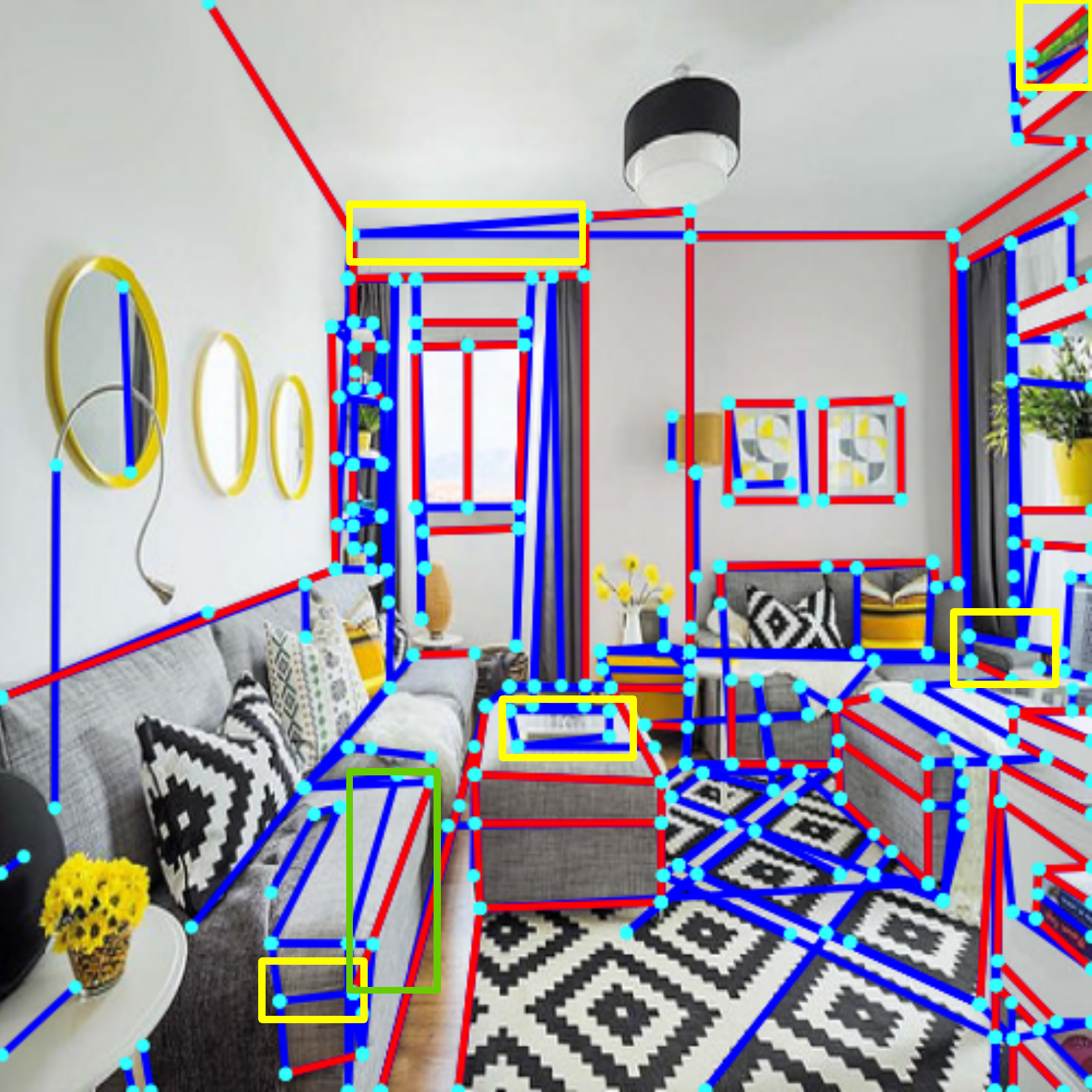}
         \caption{LAP Matching}
         \label{sub_fig:lap_match}
     \end{subfigure}

   \end{center}
    \caption{Comparison of line matching evaluation results using different metrics. (a) The ground truth line segments marked by red. (b) Line matching result using sAP$^{10}$ metric. (c) Line matching result using proposed LAP metric. In (b) and (c), the mismatched and matched line segments are marked by blue and red, respectively. The endpoints are marked by cyan.}
    \label{metric_match}
\end{figure}

Using LMS to determine true positive, i.e. a detected line segment is considered to be true positive if LMS$>0.5$, we can calculate the Line Matching Average Precision (LAP) on the entire test set.
 LAP is defined as the area under the precision recall curve. 

\textbf{Analysis of metric on real image.} We compare the line matching evaluation results between SSE used in sAP and LMS used in LAP on a real image, as shown in Fig.~\ref{metric_match}. Comparing the areas labeled by yellow boxes in Fig.~\ref{sub_fig:sap_match} and Fig.~\ref{sub_fig:gt_match}, the detected line segments have obvious error direction compared to ground truth. However, SSE gives the same tolerance for line segments with different lengths, and accepts them as true positive matches. In contrast, as shown in Fig.~\ref{sub_fig:lap_match}, LMS could better capture the direction errors and give the correct judgement. As shown by the green boxes in Fig.~\ref{sub_fig:gt_match} and Fig.~\ref{sub_fig:lap_match}, for the line segment with the correct direction but the slightly shorter length compared with the ground truth, namely, whose $Score_l$ is lower than 1 but greater than $\eta_l$, LMS will accept it while SSE would not. Considering that the direction of line segments are more important in upper-level applications such as SLAM, this deviation can be acceptable.

\section{Experiments}

Experiments are conducted on Wireframe dataset~\cite{Wireframe} and YorkUrban dataset~\cite{York}. Wireframe contains 5462 images of indoor and outdoor man-made environments, among which 5000 images are used for training. To validate the generalization ability, we also evaluate on YorkUrban Dataset~\cite{York}, which has 102 test images.

We use the standard data augmentation procedure to expand the diversity of training samples, including horizontal and vertical flip, rotation and scaling.
The hardware configuration includes four NVIDIA RTX 2080Ti GPUs and an Intel Xeon Gold 6130 2.10 GHz CPU. We use the ADAM optimizer with an initial learning rate of $1\times 10^{-3}$, which is divided by 10 at the 150th, 250th, and 350th epoch. The total training epoch is 400.

\begin{figure}[t]
    \begin{center}
     \begin{subfigure}[b]{0.19\textwidth}
         \centering
         \includegraphics[width=0.9\linewidth]{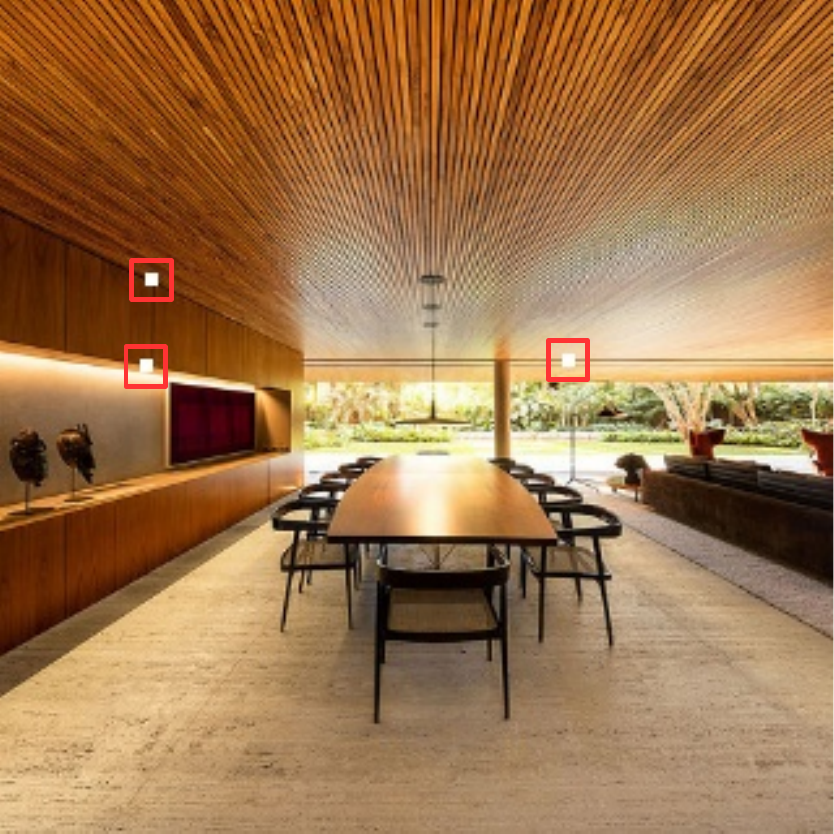}
         \caption{Image}
         \label{sub_fig:image}
     \end{subfigure}
     \begin{subfigure}[b]{0.19\textwidth}
         \centering
         \includegraphics[width=0.9\linewidth]{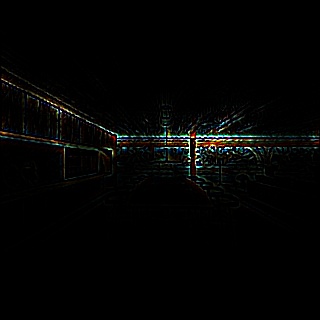}
         \caption{model 1}
         \label{sub_fig:noline}
     \end{subfigure}
     \begin{subfigure}[b]{0.19\textwidth}
         \centering
         \includegraphics[width=0.9\linewidth]{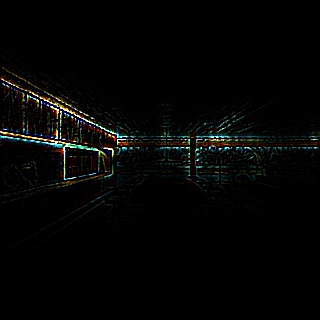}
         \caption{model 2}
         \label{sub_fig:withline}
     \end{subfigure}
     \begin{subfigure}[b]{0.19\textwidth}
         \centering
         \includegraphics[width=0.9\linewidth]{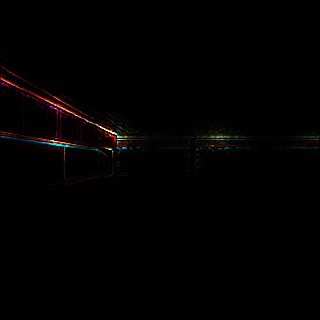}
         \caption{model 3}
         \label{sub_fig:withFAM}
     \end{subfigure}
     \begin{subfigure}[b]{0.19\textwidth}
         \centering
         \includegraphics[width=0.9\linewidth]{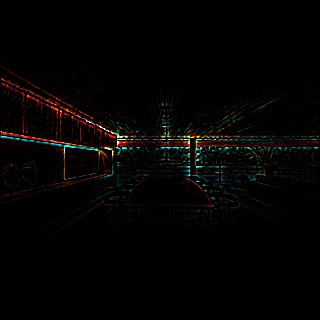}
         \caption{model 4}
         \label{sub_fig:withMixConv}
     \end{subfigure}
   \end{center}
    \caption{Gradient based interpretation of root-point detection. (a) Raw image and the root points (white dots) of three line segments. (b-e) The gradient saliency maps of the input layer backpropogated from the three root points detected by the four different models, based on Guided Back-propogation method [17].}
    \label{grad}
\end{figure}

\subsection{Analysis of TP-LSD}
\label{ablationstudy}
We run a series of ablation experiments to
study our proposed TP-LSD on Wireframe dataset. The evaluation results are shown in Table~\ref{ablation}. F$^H$ refers to pixel based metric~\cite{Wireframe}. sAP$^{10}$ is the structural average precision~\cite{LCNN} with threshold of 10. LAP is the proposed metric. As presented in Table~\ref{ablation}, all the proposed modules present contributions to the performance improvements.

\textbf{LSB.} After integrating the line-map segmentation branch with the TP extraction branch without cross-branch guidance, the multi-task learning improves the performance from 0.782 to 0.808, because the line segmentation learning can guide the model to learn more line-awareness features.

\textbf{FAM.} FAM combines the cross-branch guidance with the line-map segmentation branch. Although the F$^H$ metric increases indicating the better pixel localization accuracy, sAP$^{10}$ and LAP are slightly decreased, because of the a larger number of line segments are detected.

\textbf{MCM.} Mixture Convolution Module is applied in root-point detection sub-branch. Compared to the standard convolution layers, MCM improves the LAP scores significantly, showing a better matching degree.

\textbf{PFM.} With PFM and the contribution ratio of $\alpha=0.5$, the precision is increased while the recall slightly decreased, which lead to a better overall accuracy. The decrease in sAP$^{10}$ and LAP is due to the reduced confidence of the root-points after PFM.

\setlength{\tabcolsep}{3pt}
\begin{table}[t]
\begin{center}
\caption{Ablation study of TP-LSD on Wireframe dataset. "LSB", "FAM" and "MCM" refer to line segmentation branch, feature aggregate module and mixture convolution module, respectively. $\alpha$ is the contribution ratio of line-map as Eq.~\eqref{combine}. "R/S" refers to the rotation and scale data augmentation strategy. "Avg. line Num." means the average number of detected line segments whose confidence are greater than $0.2$.}
\label{ablation}
\begin{tabular}{c|ccccc|ccc| c}
No. & LSB & FAM & MCM & PFM & R/S & F$^H$ & sAP$^{10}$ & LAP & Avg. line\\
\hline
\hline
1 & & & &-& \checkmark &0.782 & 56.8 &58.6 & 116.1  \\
2 & \checkmark  &  & &-& \checkmark &0.808 &59.2 &61.4 &113.6\\
3 & \checkmark & \checkmark & &-& \checkmark &0.810 &60.4 &59.2 & 138.0\\
4 & \checkmark & \checkmark &\checkmark &0&\checkmark & 0.810  &61.3  & 60.9& /  \\
5 &\checkmark&\checkmark&\checkmark &1&\checkmark & 0.811  &60.0  &60.1 &/  \\
6 &\checkmark&\checkmark&\checkmark&0.5 &\checkmark& 0.816  &60.6  &60.6 &/  \\
7 & \checkmark & \checkmark &\checkmark &0.5 &$\times$ & 0.813  &60.0 &60.1 &/ \\
\hline
\end{tabular}
\end{center}
\end{table}

\textbf{Augmentation.} The $7^{th}$ row in Table~\ref{ablation} shows the data augmentation with only horizontal and vertical flip. Compared to the result in $6^{th}$ row, the lower performance shows that the rotation and scaling based data augmentation can further improve the performance.

\textbf{Interpretability.} To explore what the network learned from the line segment detection task, we use Guided Backpropogation~\cite{backprop} to visualize which pixels are important for the root-point detection. Guided Backpropogation interprets the pixels' importance degree on the input image, by calculating the gradient flow from the output layer to the input images. 
The gradients flowed to the input images from the three specific detected root-point are visualized in Fig.~\ref{grad}. We find that the network automatically learns to localize the saliency region w.r.t a root-point, which is along a complete line segment. It shows that the root point detection task is mainly based on the line feature.

Furthermore, the integration of LSB lead to the higher influence of on-line pixels to root point prediction. Comparing Fig. \ref{sub_fig:withline} to Fig.~\ref{sub_fig:noline}, the former presents higher gradient values along the line. The saliency maps obtained by model No. 3 and model No. 4 are cleaner, and the saliency regions are more concentrated on specific line segments. With the introduction of MCM in model No. 4, the response of long line segment could be improved with a lager receptive field, which can be shown by the comparison between Fig.~\ref{sub_fig:withFAM} and Fig.~\ref{sub_fig:withMixConv}.

\setcounter{footnote}{0}
\subsection{Comparison with other methods}

We compare our proposed TP-LSD with LSD\footnote{\url{http://www.ipol.im/pub/art/2012/gjmr-lsd/}}~\cite{LSD}, DWP\footnote{\url{https://github.com/huangkuns/wireframe}}~\cite{Wireframe}, AFM\footnote{\url{https://github.com/cherubicXN/afm-cvpr2019}}~\cite{AFM}, L-CNN\footnote{\url{https://github.com/zhou13/lcnn}} and L-CNN with post-process (L-CNN(P))~\cite{LCNN}. The source codes and their model weights provided by the authors are available online, except that we reproduced DWP by ourselves. F$^H$, sAP and LAP are used to evaluate those methods quantitatively.
For TP-LSD, we tried a series of minimum thresholds of the root-point detection confidence, ranging within (0.1, 0.8) with the step $\triangle \gamma = 0.05$. LSD is evaluated with $-\log($NFA$)$ in $0.01 \times \{1.75^0, ..., 1.75^{19}\}$, where NFA is the number of false positive detections. For other methods, we use the author recommended threshold array listed in \cite{Wireframe, AFM, LCNN}.

\begin{table}[t]
\small
\begin{center}
\setlength{\tabcolsep}{1mm}
\caption{Evaluation results of different line segment detection methods. "/" means that the score is too slow to be meaningful. The best two scores are shown in \textcolor{red}{red} and \textcolor{blue}{blue}.}
\label{eval_table}
\newcommand{\tabincell}[2]{\begin{tabular}{@{}#1@{}}#2\end{tabular}}
\begin{tabular}{l c cccc cccc c }
\noalign{\smallskip}
\hline
\noalign{\smallskip}
\multirow{2}*{Method} 
& \multirow{2}*{\shortstack{Input \\Size}}  
& \multicolumn{4}{c}{Wireframe dataset}
& \multicolumn{4}{c}{YorkUrban dataset}  
& \multirow{2}*{FPS} \\

\cmidrule(r){3-6} \cmidrule(r){7-10}
&   &  F$^H$      &  sAP$^{5}$ & sAP$^{10}$  &  LAP
    &  F$^H$     &  sAP$^{5}$ &  sAP$^{10}$  & LAP \\

\hline
LSD~\cite{LSD}   & $320$  &0.641    &6.7&8.8    &18.7
& 0.606 & 7.5&9.2 & 16.1
& \textcolor{red}{100}  \\

DWP~\cite{Wireframe}  & $512$ & 0.727    &/&/    &6.6
& 0.652 &/& / & 3.1
& 2.2 \\

AFM~\cite{AFM}& $320$  &0.773    &18.3&23.9   &36.7
& 0.663  & 7.0&9.1 & 17.5
& 12.8  \\

L-CNN~\cite{LCNN}  & $512$  &0.775  &\textcolor{red}{58.9}  &\textcolor{red}{62.8}  &59.8
& 0.646 &\textcolor{blue}{25.9}& \textcolor{red}{28.2} &\textcolor{blue}{32.0}
& 11.1  \\
L-CNN(P)~\cite{LCNN}  & $512$  &\textcolor{blue}{0.817}   &52.4&57.3   &57.9
&  \textcolor{blue}{0.675} & 20.9&23.1 &  26.8
& 5.2  \\
\hline
TP-LSD-Lite & $320$ &  0.804 &56.4& 59.7 & 59.7
& \textcolor{red}{0.681} & 24.8 &26.8 & 31.2
& \textcolor{red}{78.2}
  \\
TP-LSD-Res34 & $320$ & 0.816 &57.5 &\textcolor{blue}{60.6}  &\textcolor{blue}{60.6}
&0.674 &25.3 &27.4  &31.1
&42.2
  \\
TP-LSD-HG & $512$ & \textcolor{red}{0.820}  &50.9 &57.0 &55.1
&0.673  &18.9  &22.0 & 24.6
&\textcolor{blue}{  53.4}
  \\
TP-LSD-Res34 &$512$   & 0.806  & \textcolor{blue}{57.6}  & 57.2  &  \textcolor{red}{61.3} &  0.672 &  \textcolor{red}{27.6} & \textcolor{blue}{27.7}& \textcolor{red}{34.3}  &18.1
\\
\hline
\end{tabular}
\end{center}
\end{table}

\begin{figure}[t]
    \begin{center}
     \begin{subfigure}[b]{0.24\textwidth}
         \centering
         \includegraphics[width=0.95\linewidth]{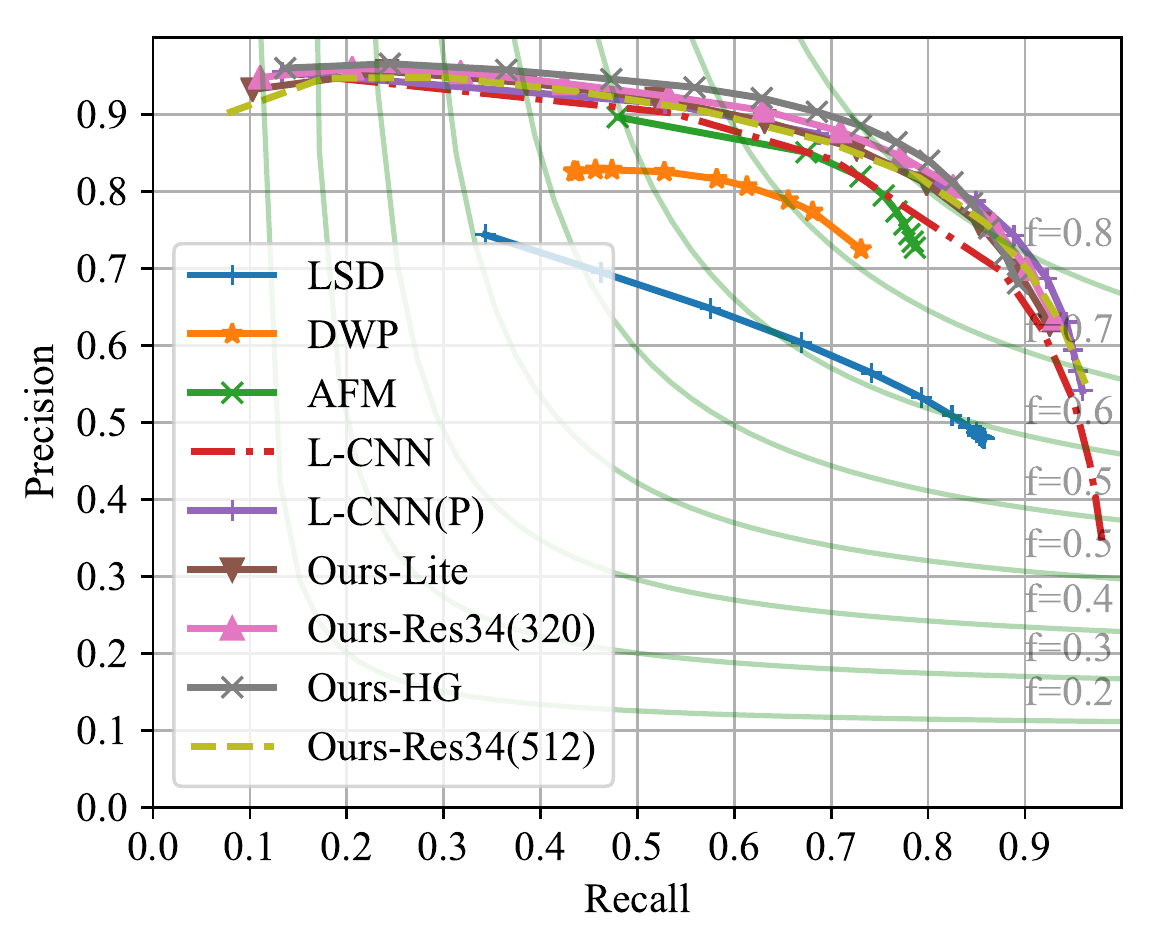}
         \caption{\centering Pixel based metric; Wireframe} 
         \label{sub_fig:pr_a}
     \end{subfigure}
     \begin{subfigure}[b]{0.24\textwidth}
         \centering
         \includegraphics[width=0.95\linewidth]{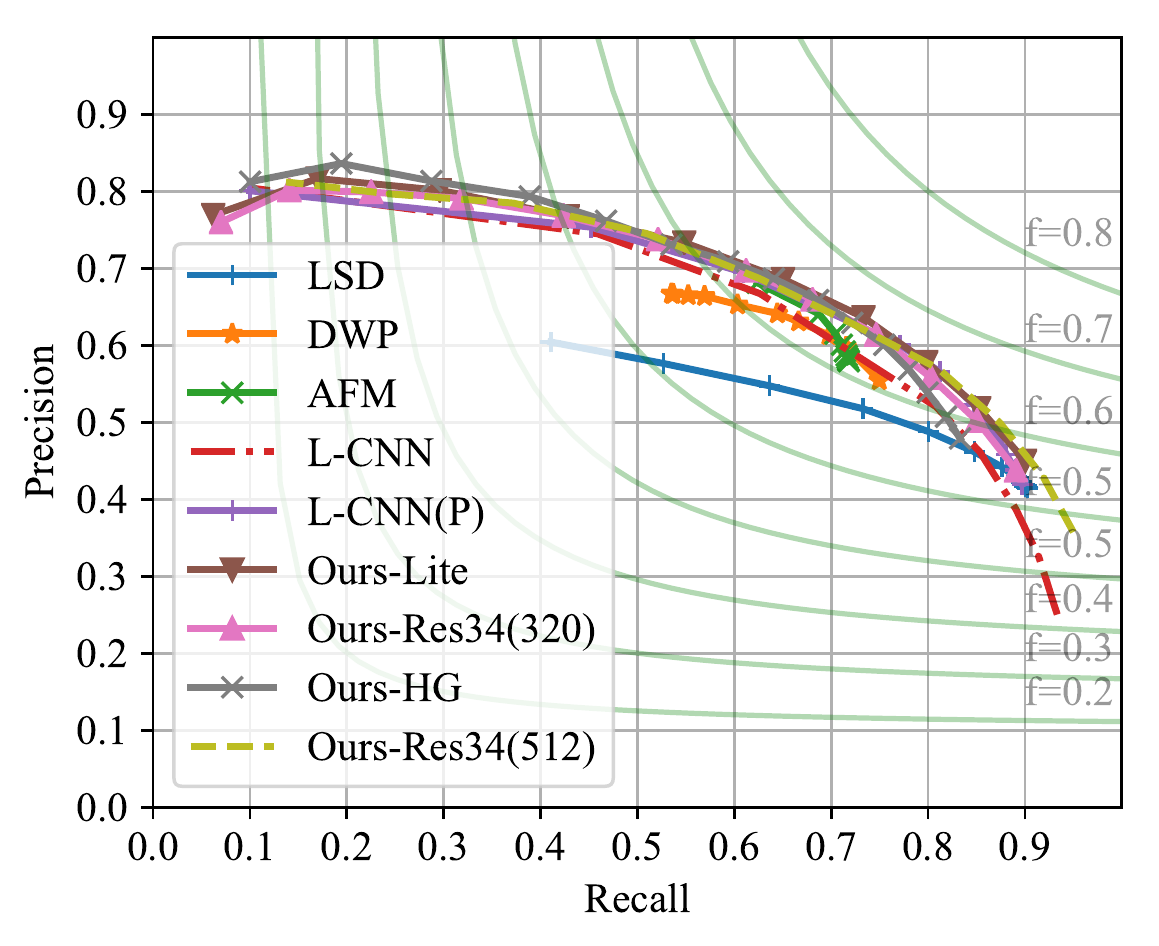}
         \caption{\centering Pixel based metric; YorkUrban} 
         \label{sub_fig:pr_b}
     \end{subfigure}
     \begin{subfigure}[b]{0.24\textwidth}
         \centering
         \includegraphics[width=0.95\linewidth]{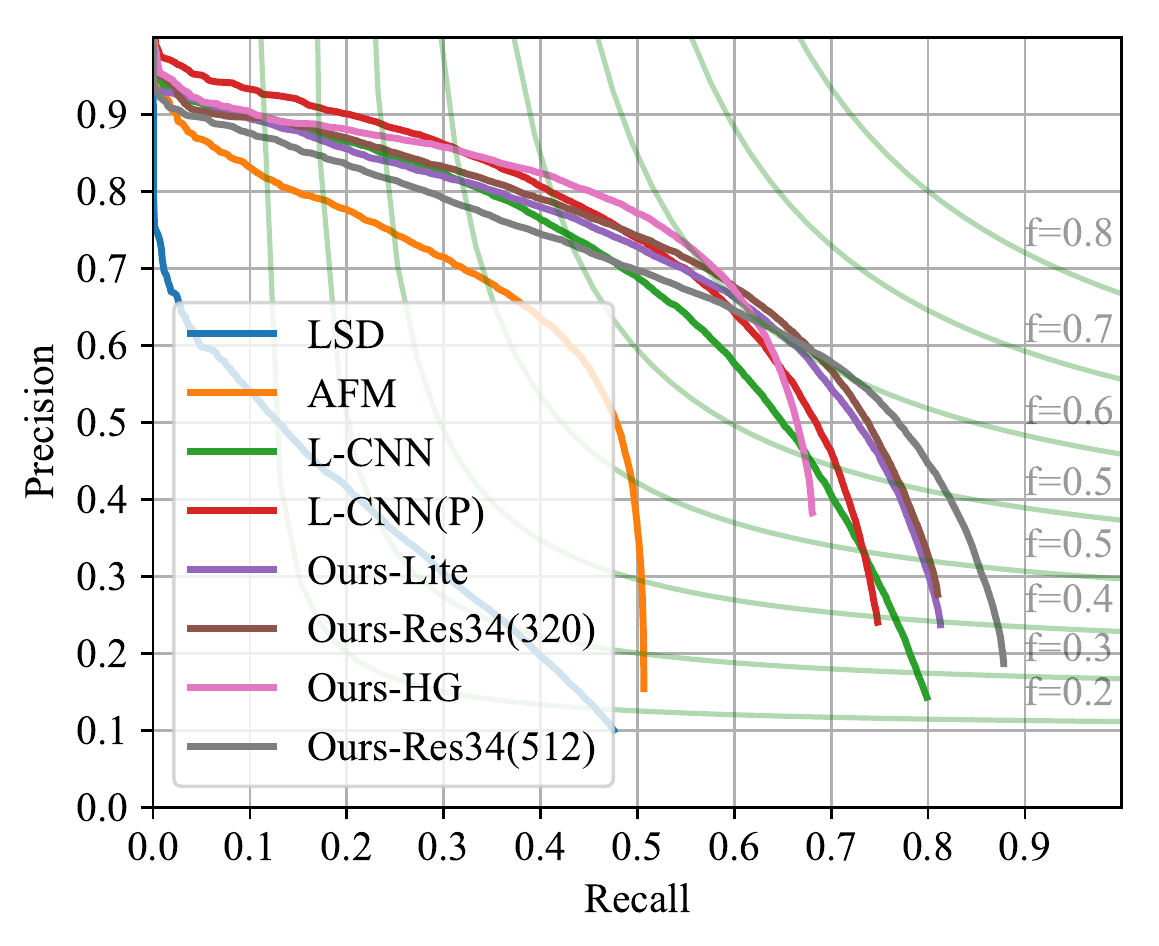}
         \caption{\centering LMS metric; Wireframe} %
         \label{sub_fig:pr_c}
     \end{subfigure}
     \begin{subfigure}[b]{0.24\textwidth}
         \centering
         \includegraphics[width=0.95\linewidth]{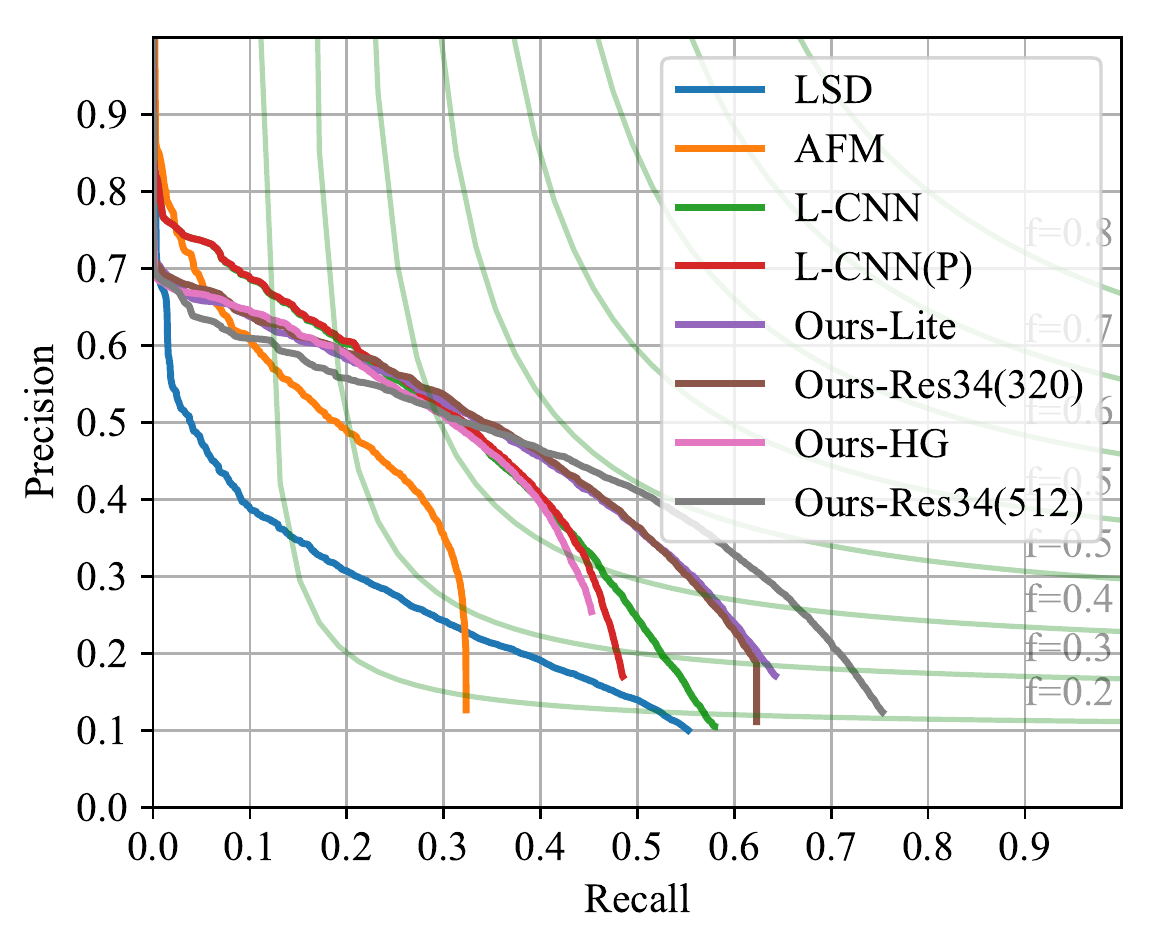}
         \caption{\centering LMS metric; YorkUrban} 
         \label{sub_fig:pr_d}
     \end{subfigure}
  \end{center}
    \caption{Precision-recall curves of line segment detection. The models are trained on Wireframe dataset and tested on both Wireframe and YorkUrban datasets. Scores below $0.1$ are not plotted. The PR curves for LAP of DWP are not ploted for its lower score.}
    \label{PR_pixel}
\end{figure}

We evaluate the methods on Wireframe and YorkUrban dataset. We use the model No. 6 in Section~\ref{ablationstudy} as the representative model, named as TP-LSD-Res34. Furthermore, we alter the backbone with Hourglass used in L-CNN~\cite{LCNN} to form TP-LSD-HG. To achieve a faster speed, TP-LSD-Lite is realized by using the output of the last second layer of the decoder as the shared feature. Thus the input to the task branches has the smaller size of 160$\times$160. And the final output of the task branches are upsampled back to 320$\times$320 with the bi-linear interpolation.

The precision-recall curves are depicted in Fig.~\ref{PR_pixel} and the detection performances are reported in Table \ref{eval_table}.
Fig.~\ref{sub_fig:pr_a} and Fig.~\ref{sub_fig:pr_b} show that TP-LSD outperforms other line segment detection methods, according to the pixel based PR curves. In addition, our one-step method provides the comparable detection performance compared to the two-step L-CNN that requires post-processing.

We then evaluate the methods with the sAP and the proposed LAP. The precision recall curve of LAP in two datasets are drawn in Fig.~\ref{sub_fig:pr_c} and Fig.~\ref{sub_fig:pr_d}. The performances of AFM and LSD are limited by length prediction of line segments.
As to DWP, the inaccurate direction prediction might affect the detection.
Our method and L-CNN present the higher scores, which shows that these two methods perform better not only in detection but also in alignment. Moreover, our method has the better precision than L-CNN in higher recall region.
Though the higher F$^H$ is obtained by TP-LSD-HG, the decreases of sAP and LAP were caused by the lower recall rate due to the lower feature map resolution. TP-LSD-Lite gets comparable generalization performance on both dataset.
YorkUrban dataset is more challenging because only the line segments which satisfy the Manhattan World assumption are labeled out as ground truth, which causes lower precision.

\begin{figure}[H]
     \begin{center}
     \begin{subfigure}[b]{0.15\textwidth}
         \centering
         \includegraphics[width=1\linewidth]{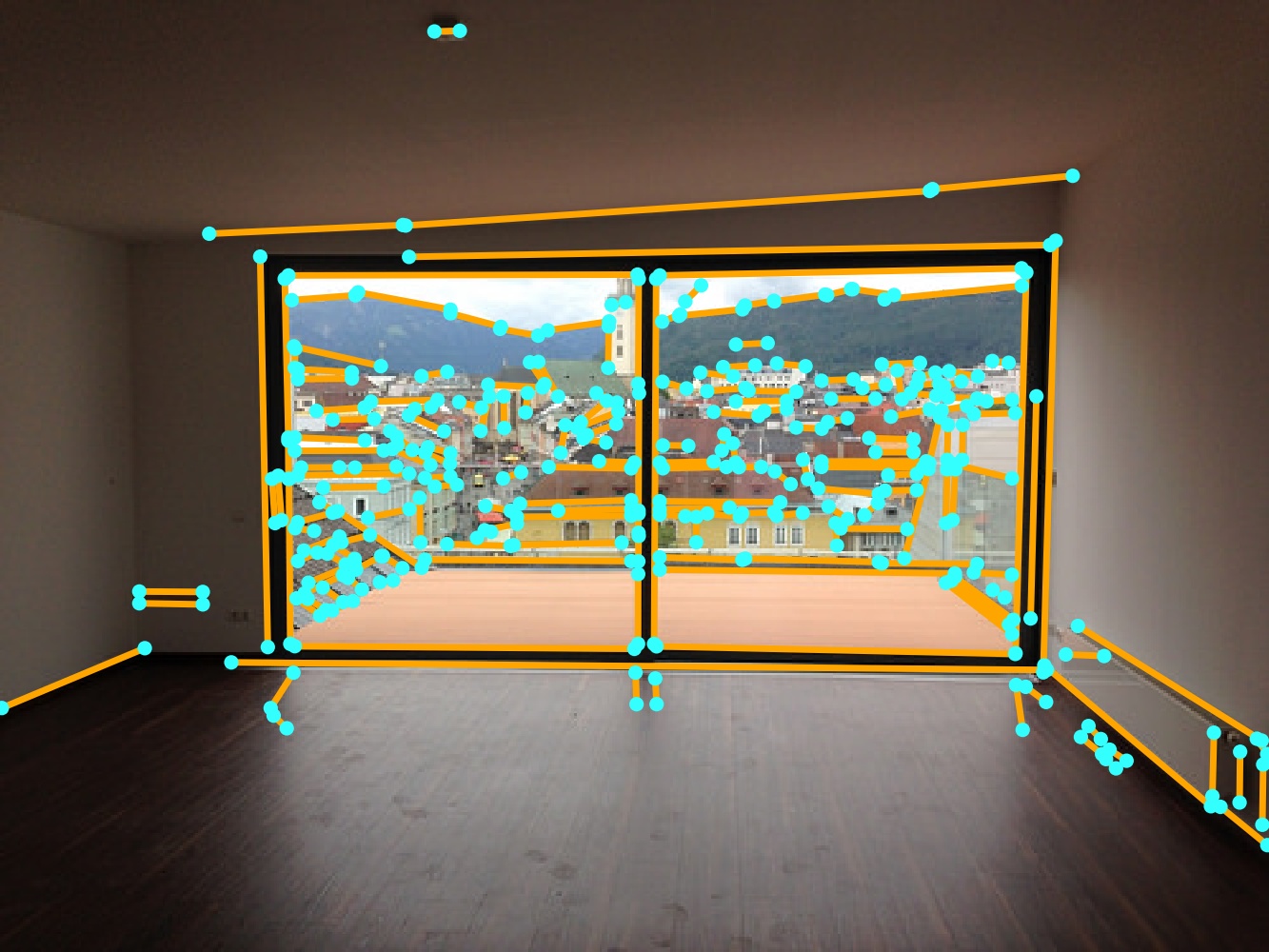}
     \end{subfigure}
    \begin{subfigure}[b]{0.15\textwidth}
         \centering
         \includegraphics[width=1\linewidth]{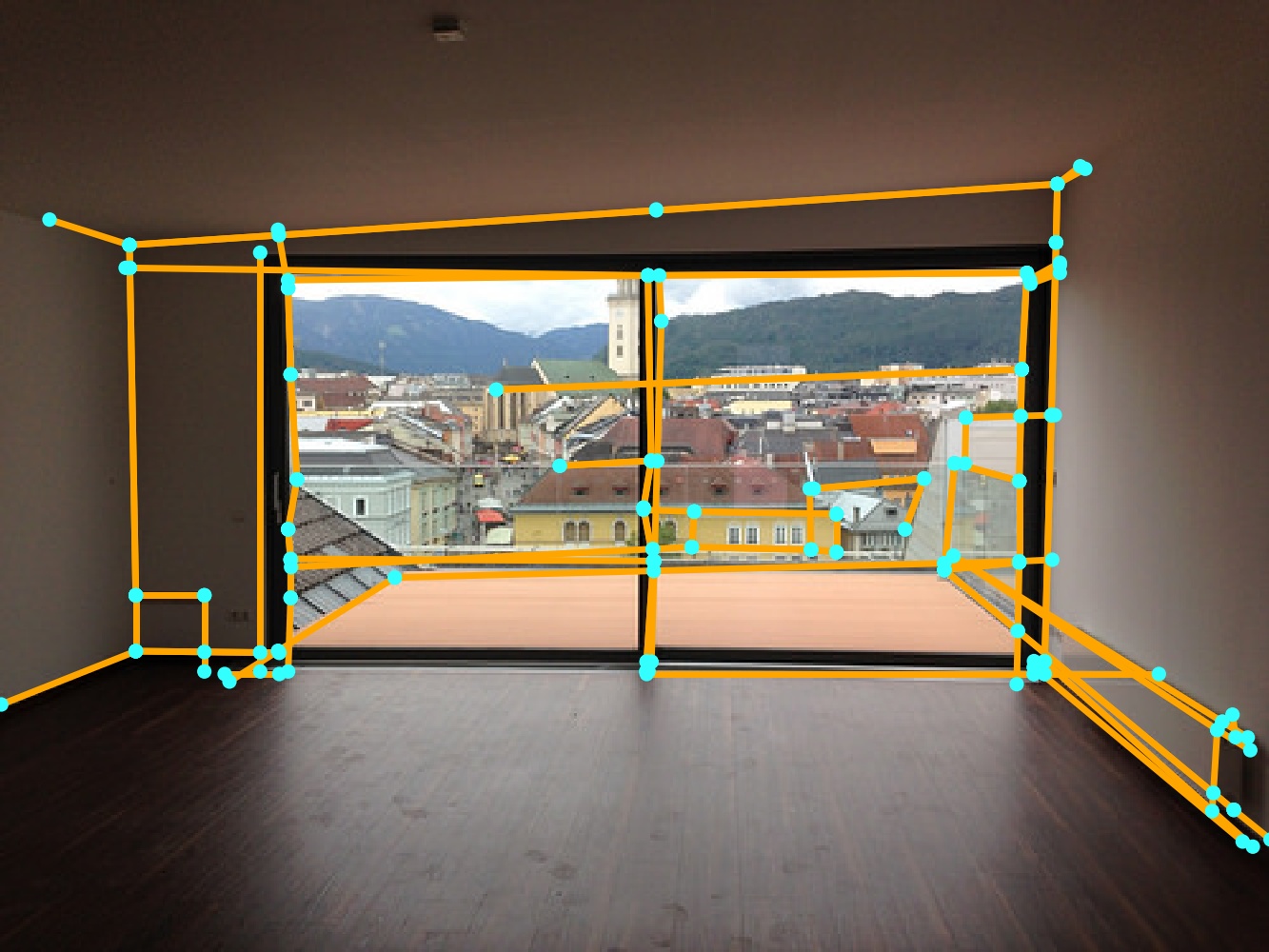}
     \end{subfigure}
     \begin{subfigure}[b]{0.15\textwidth}
         \centering
         \includegraphics[width=1\linewidth]{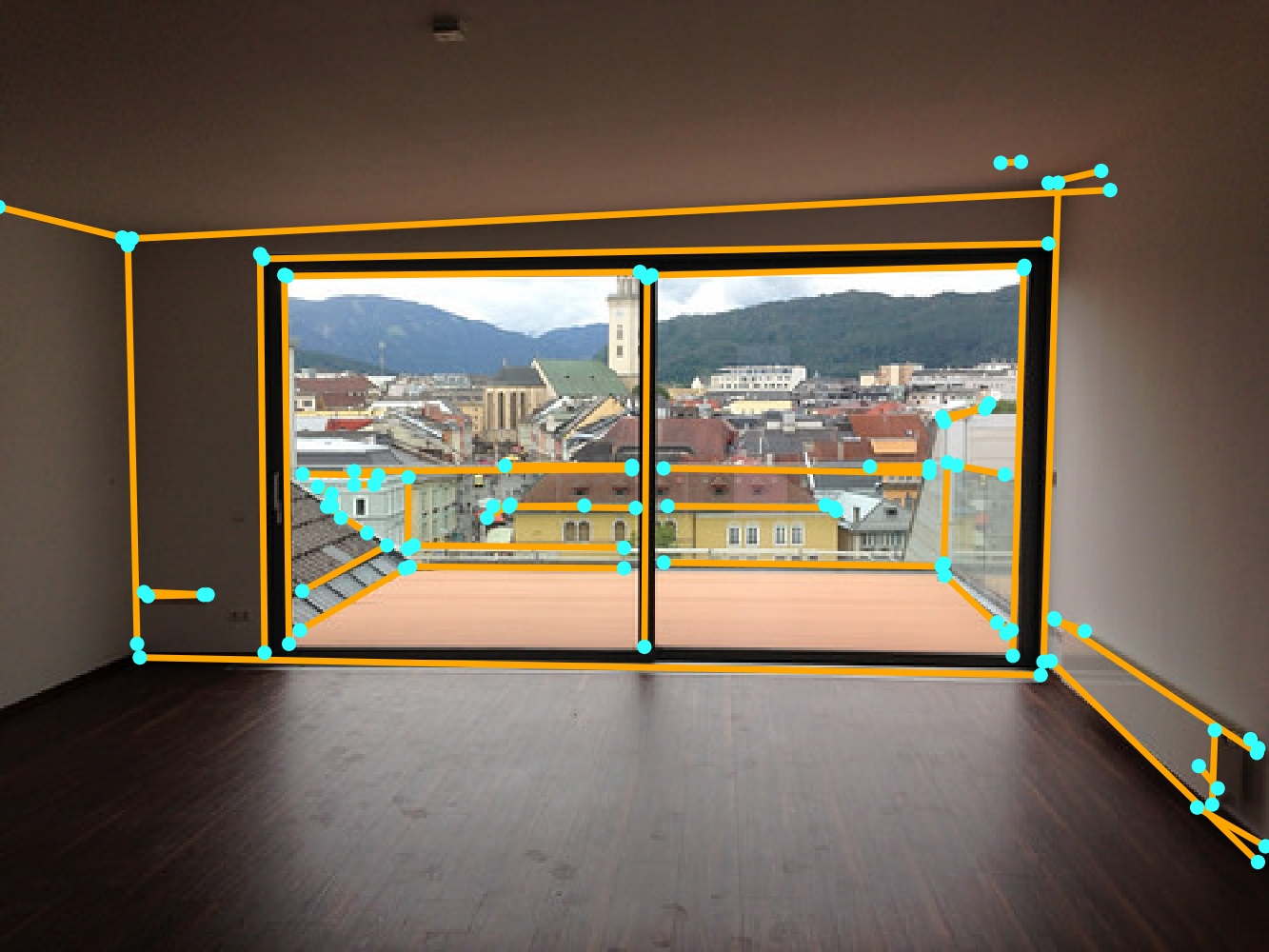}
     \end{subfigure}
     \begin{subfigure}[b]{0.15\textwidth}
         \centering
         \includegraphics[width=1\linewidth]{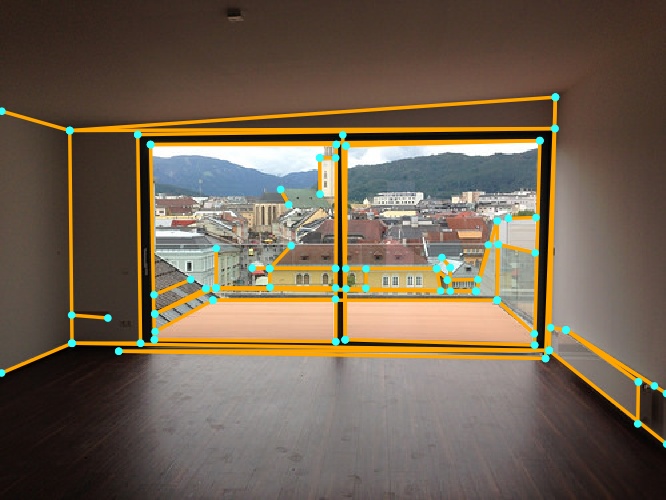}
     \end{subfigure}
     \begin{subfigure}[b]{0.15\textwidth}
         \centering
         \includegraphics[width=1\linewidth]{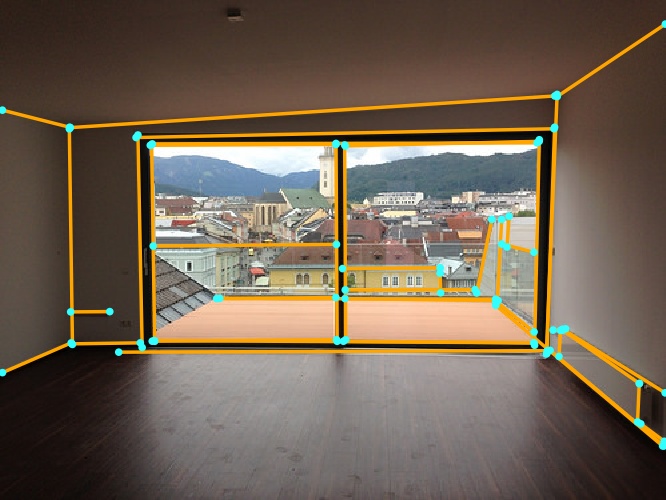}
     \end{subfigure}
     \begin{subfigure}[b]{0.15\textwidth}
         \centering
         \includegraphics[width=1\linewidth]{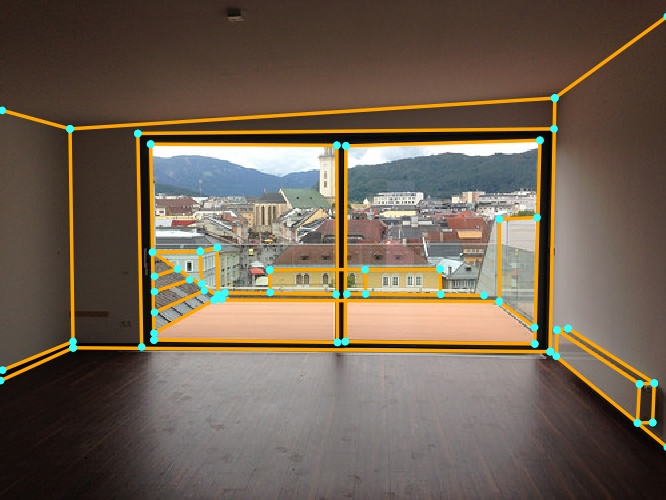}
     \end{subfigure}
     ~~
     
      \begin{subfigure}[b]{0.15\textwidth}
         \centering
         \includegraphics[width=1\linewidth]{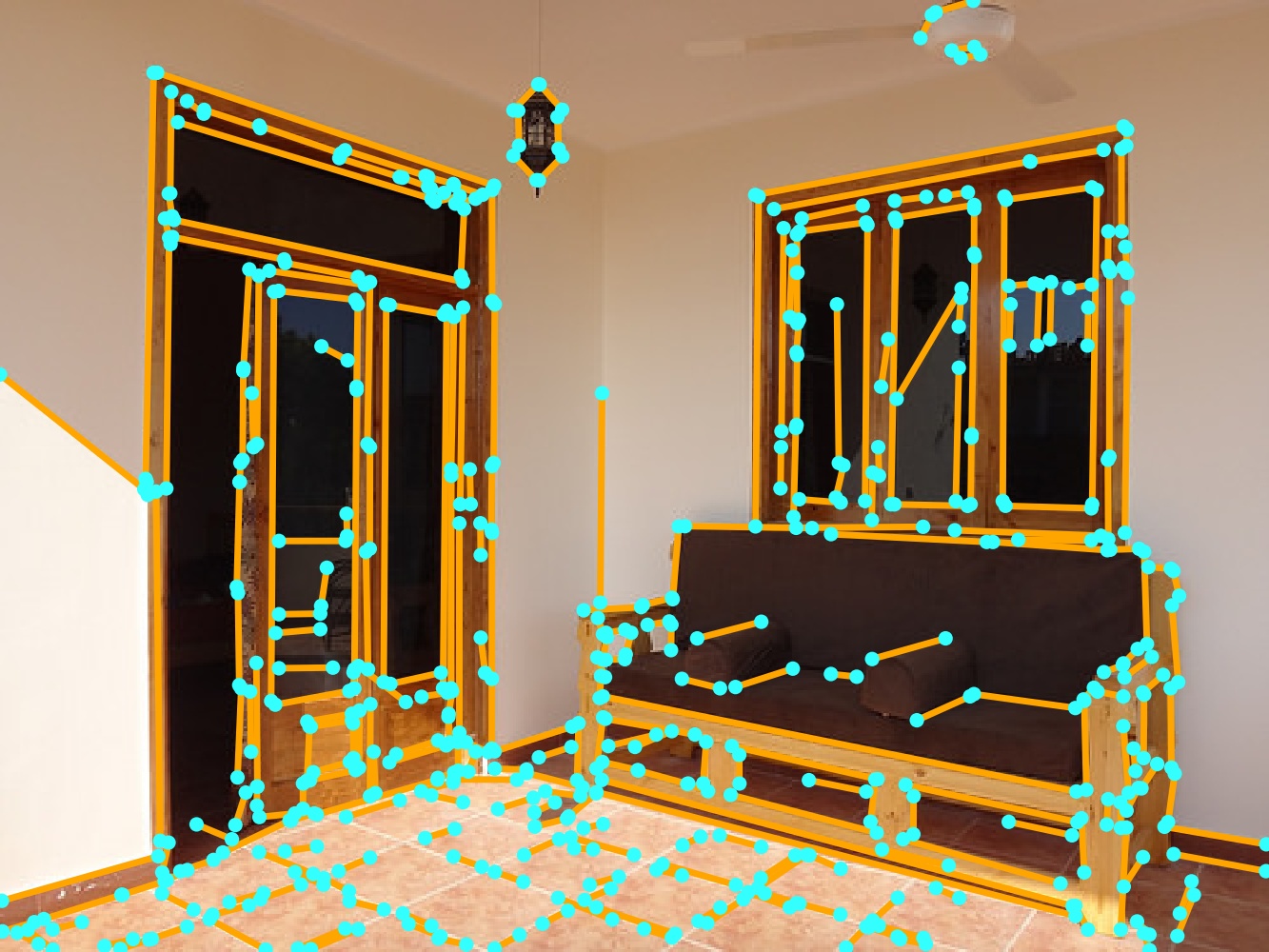}
     \end{subfigure}
    \begin{subfigure}[b]{0.15\textwidth}
         \centering
         \includegraphics[width=1\linewidth]{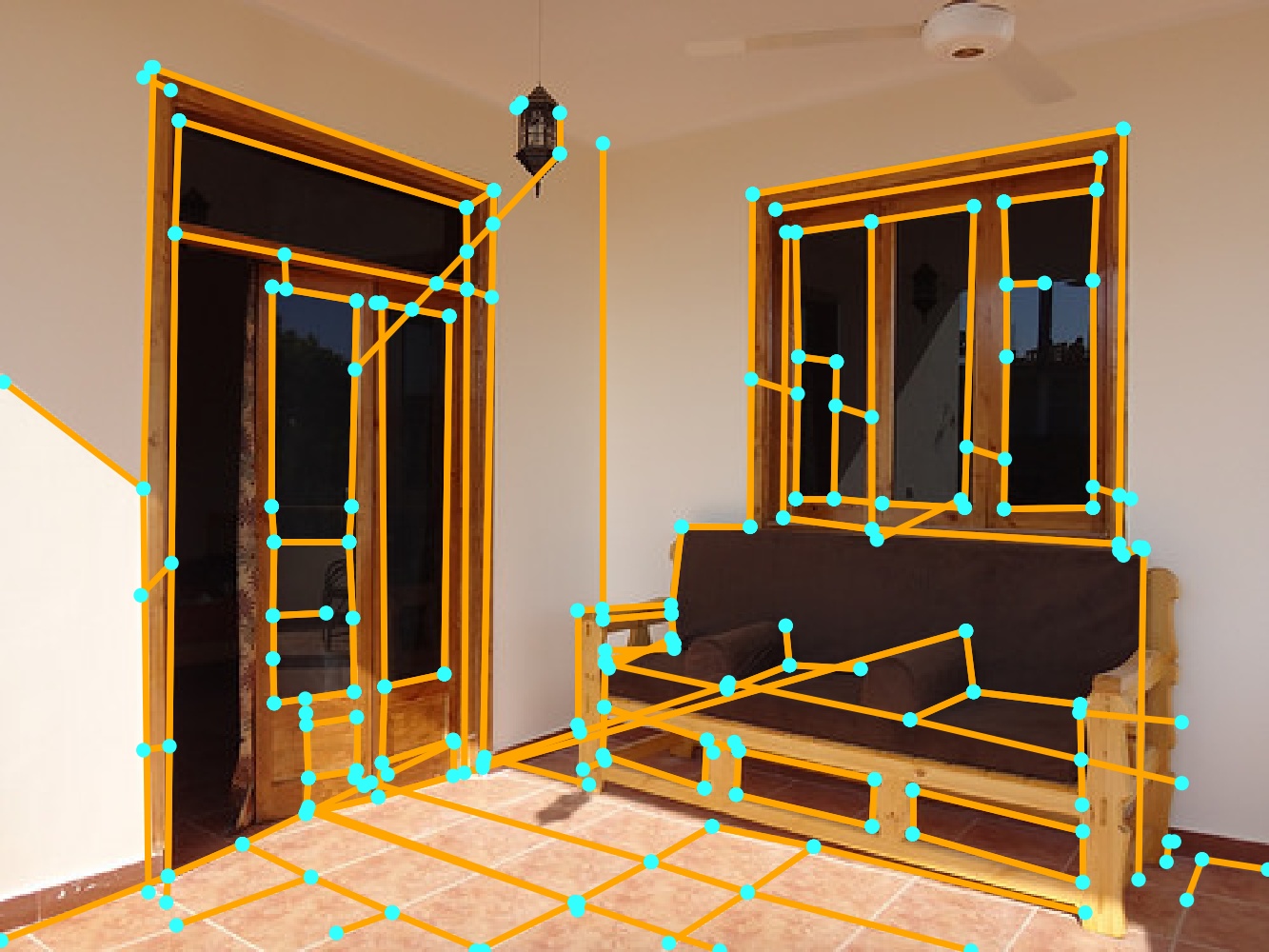}
     \end{subfigure}
     \begin{subfigure}[b]{0.15\textwidth}
         \centering
         \includegraphics[width=1\linewidth]{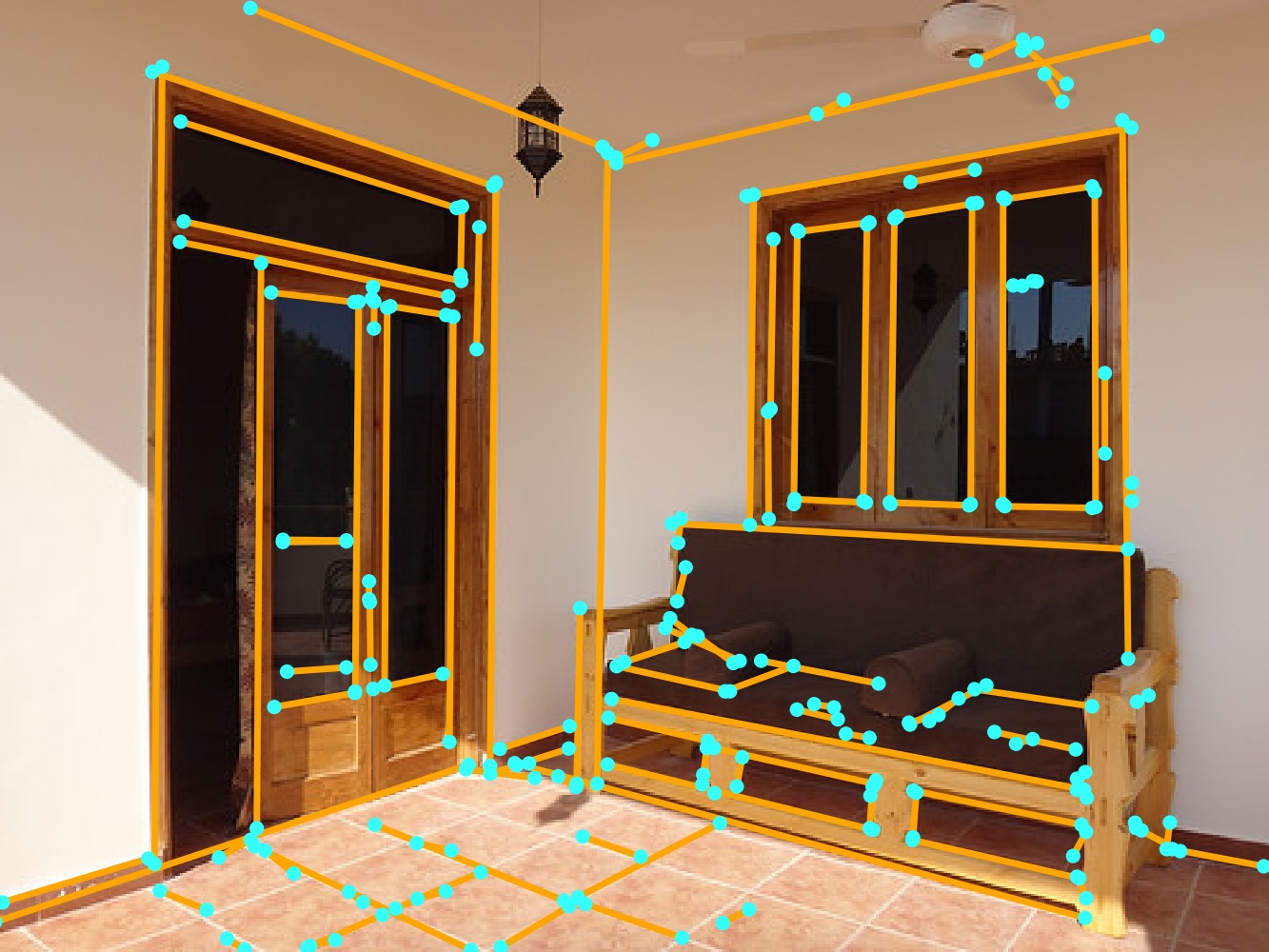}
     \end{subfigure}
     \begin{subfigure}[b]{0.15\textwidth}
         \centering
         \includegraphics[width=1\linewidth]{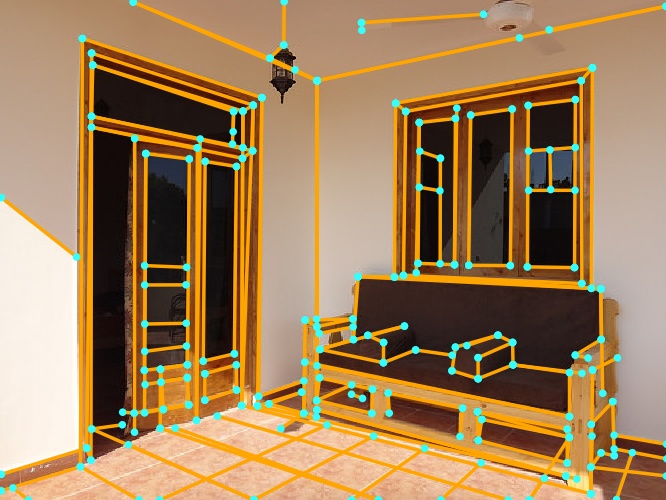}
     \end{subfigure}
     \begin{subfigure}[b]{0.15\textwidth}
         \centering
         \includegraphics[width=1\linewidth]{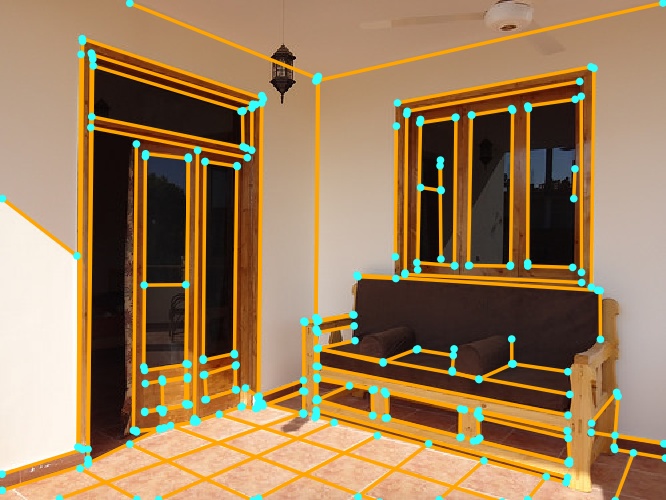}
     \end{subfigure}
     \begin{subfigure}[b]{0.15\textwidth}
         \centering
         \includegraphics[width=1\linewidth]{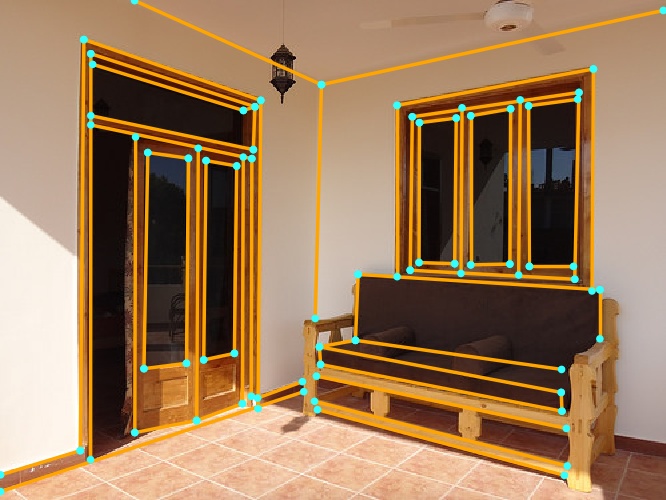}
     \end{subfigure}
     ~~
     
     \begin{subfigure}[b]{0.15\textwidth}
         \centering
         \includegraphics[width=1\linewidth]{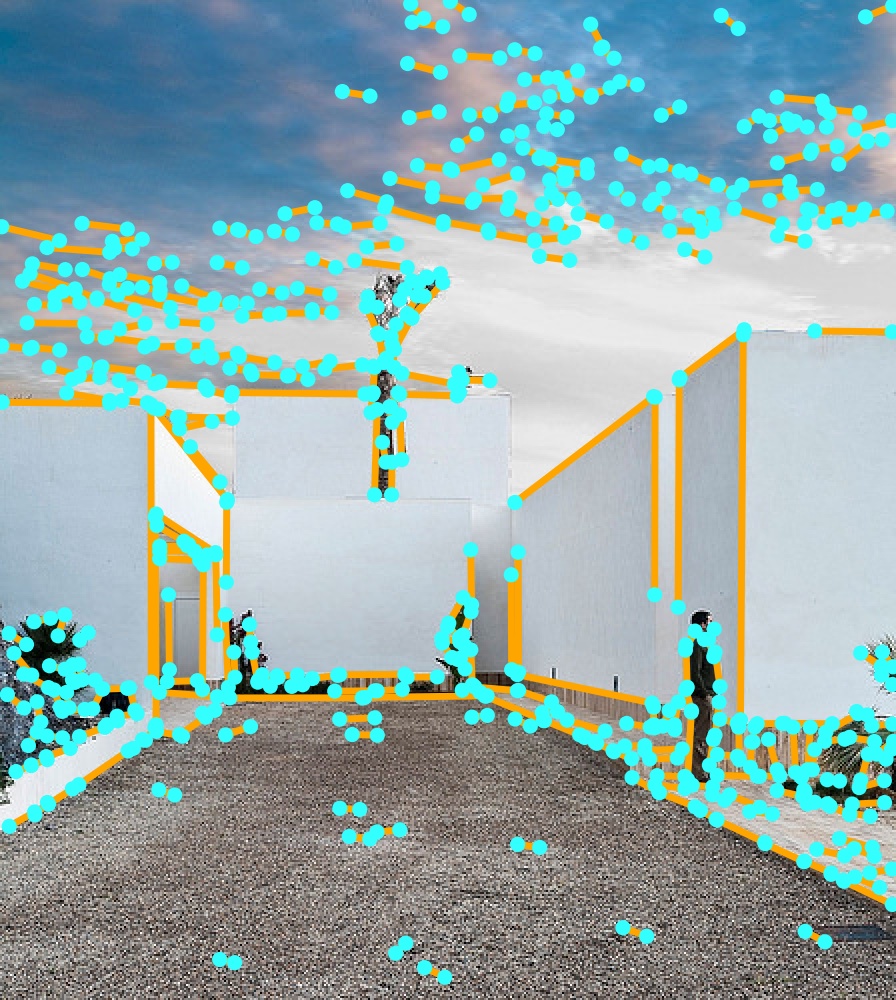}
     \end{subfigure}
    \begin{subfigure}[b]{0.15\textwidth}
         \centering
         \includegraphics[width=1\linewidth]{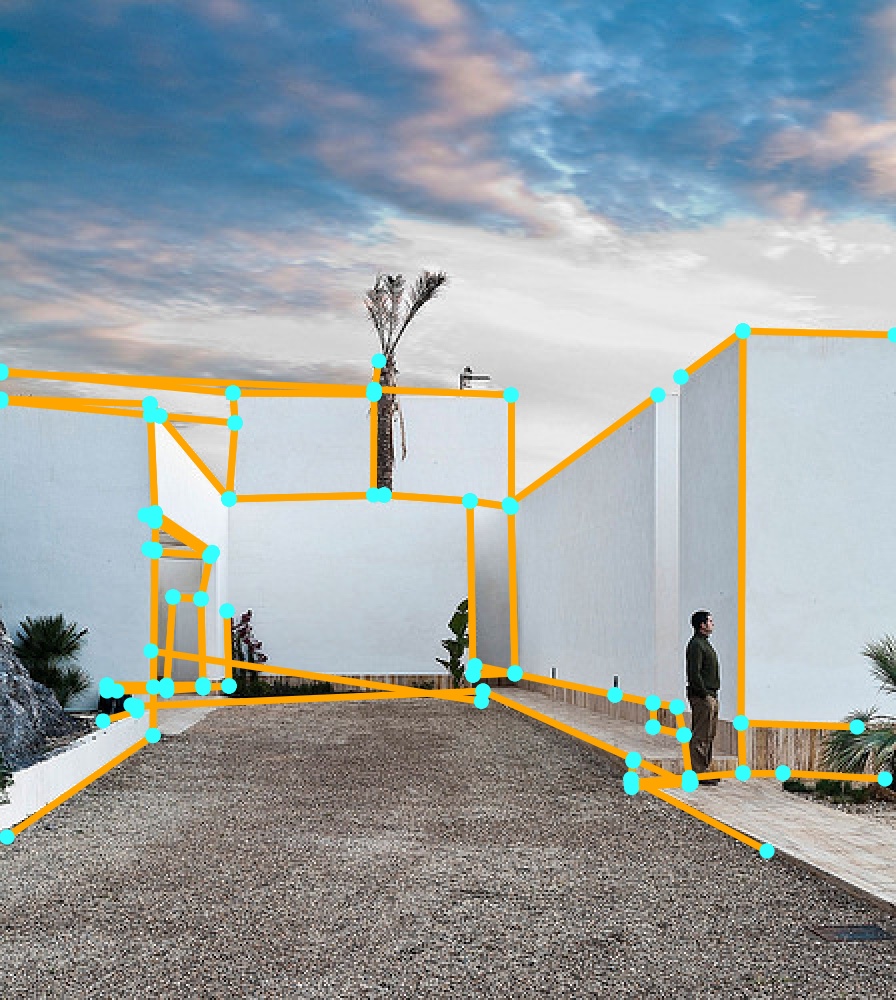}
     \end{subfigure}
     \begin{subfigure}[b]{0.15\textwidth}
         \centering
         \includegraphics[width=1\linewidth]{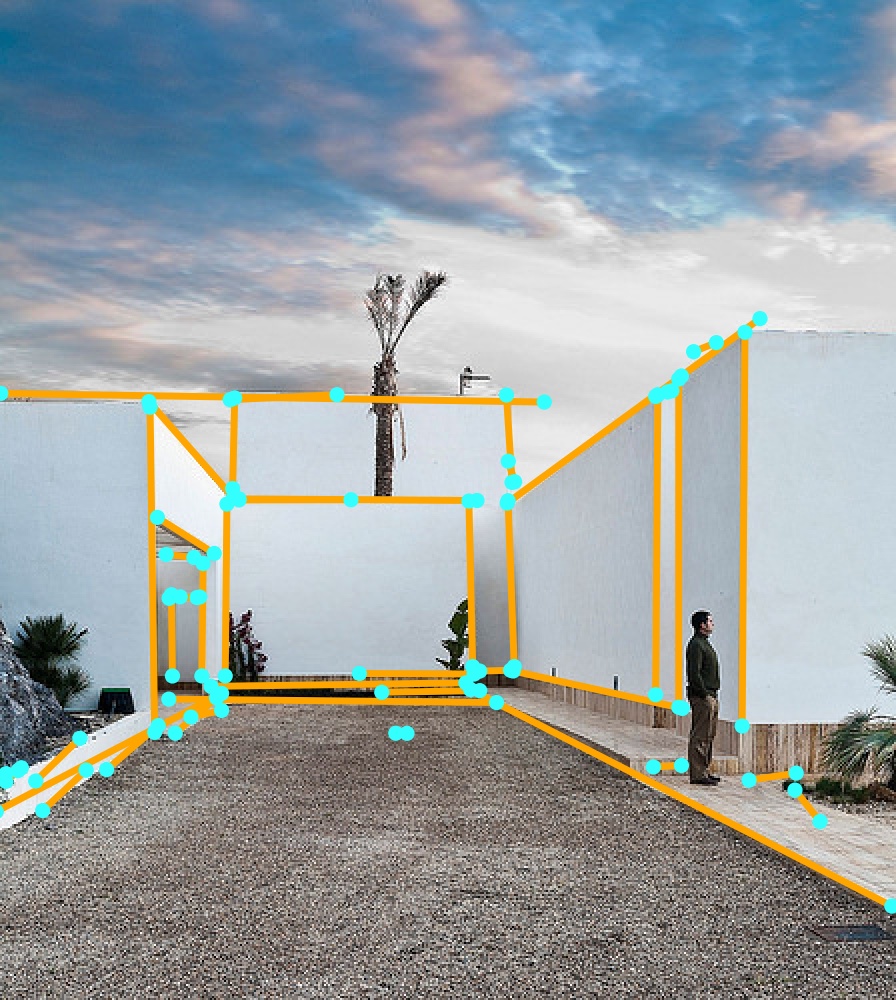}
     \end{subfigure}
     \begin{subfigure}[b]{0.15\textwidth}
         \centering
         \includegraphics[width=1\linewidth]{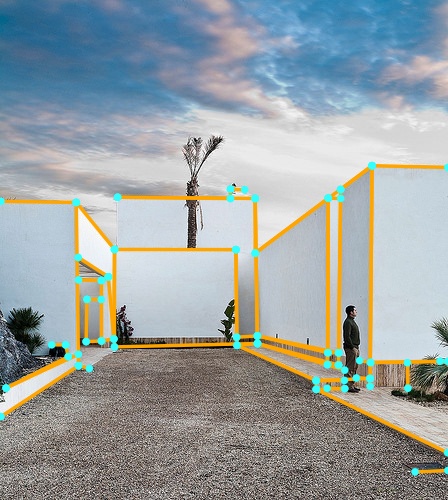}
     \end{subfigure}
     \begin{subfigure}[b]{0.15\textwidth}
         \centering
         \includegraphics[width=1\linewidth]{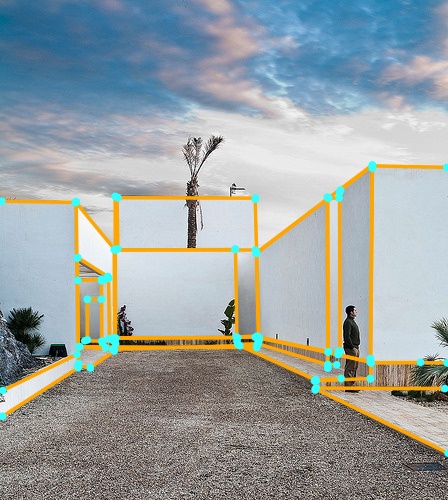}
     \end{subfigure}
     \begin{subfigure}[b]{0.15\textwidth}
         \centering
         \includegraphics[width=1\linewidth]{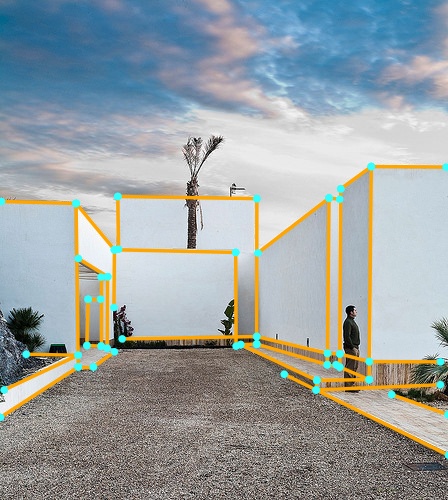}
     \end{subfigure}
     ~~
     
     \begin{subfigure}[b]{0.15\textwidth}
         \centering
         \includegraphics[width=1\linewidth]{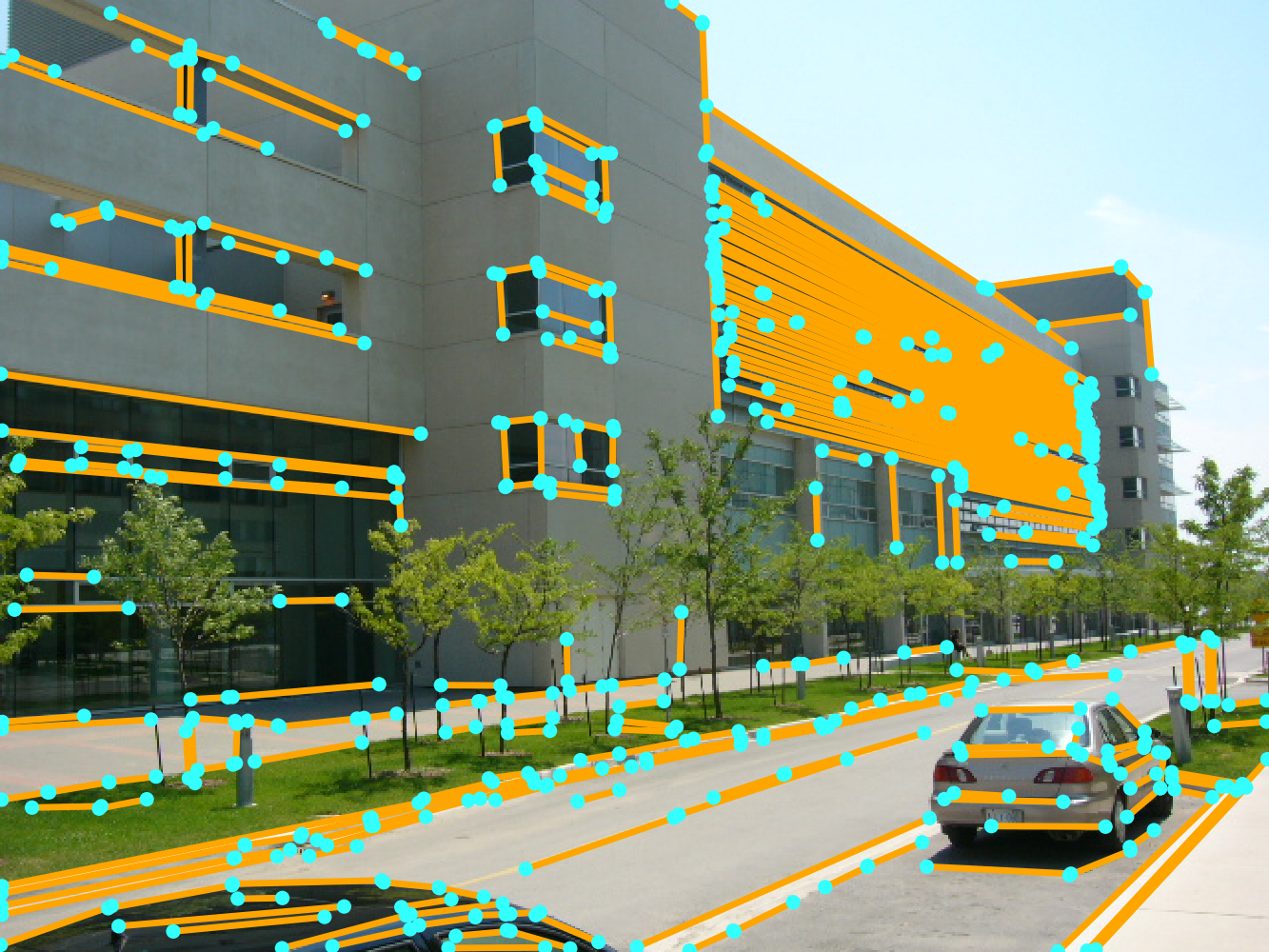}
     \end{subfigure}
    \begin{subfigure}[b]{0.15\textwidth}
         \centering
         \includegraphics[width=1\linewidth]{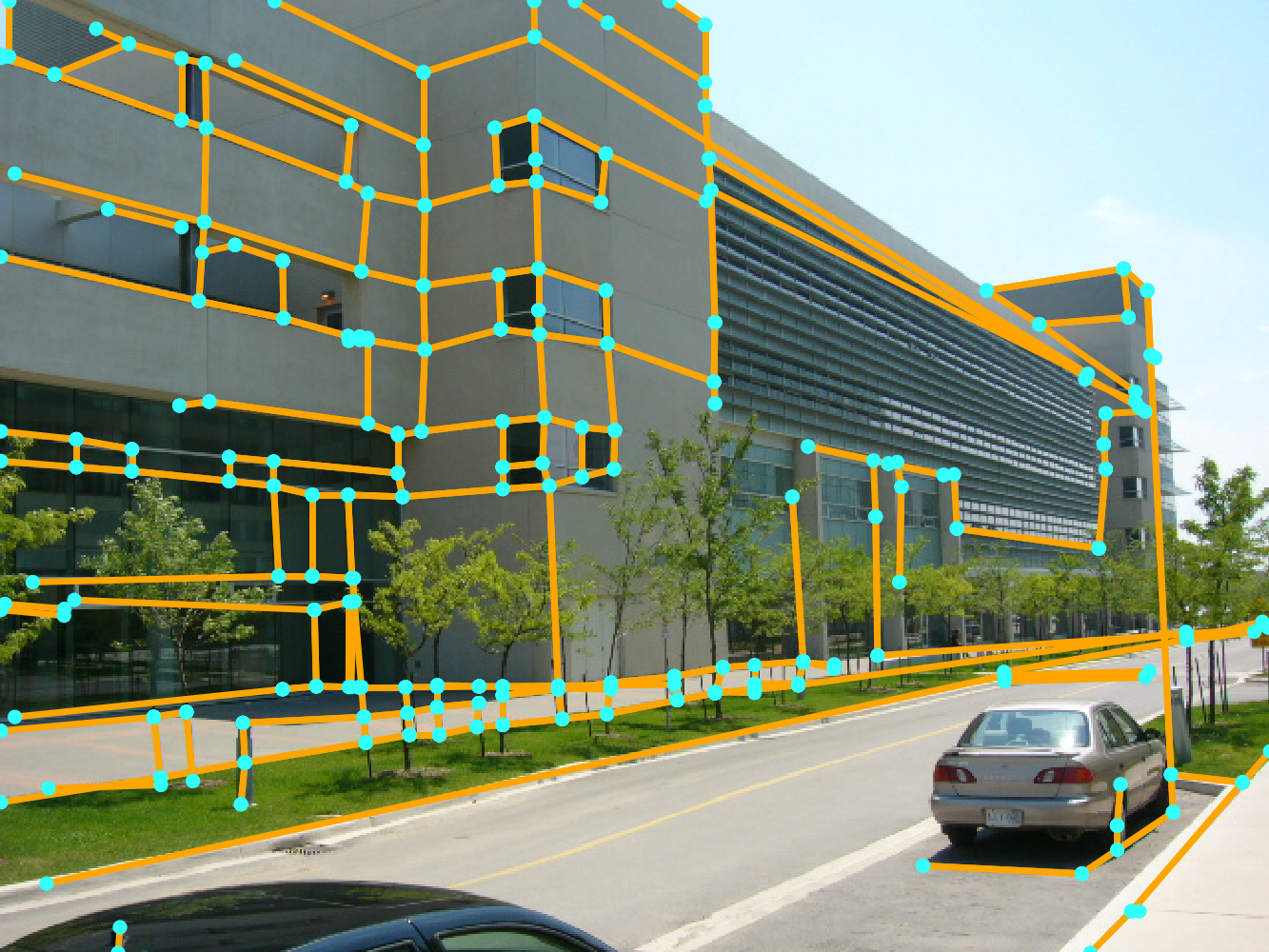}
     \end{subfigure}
     \begin{subfigure}[b]{0.15\textwidth}
         \centering
         \includegraphics[width=1\linewidth]{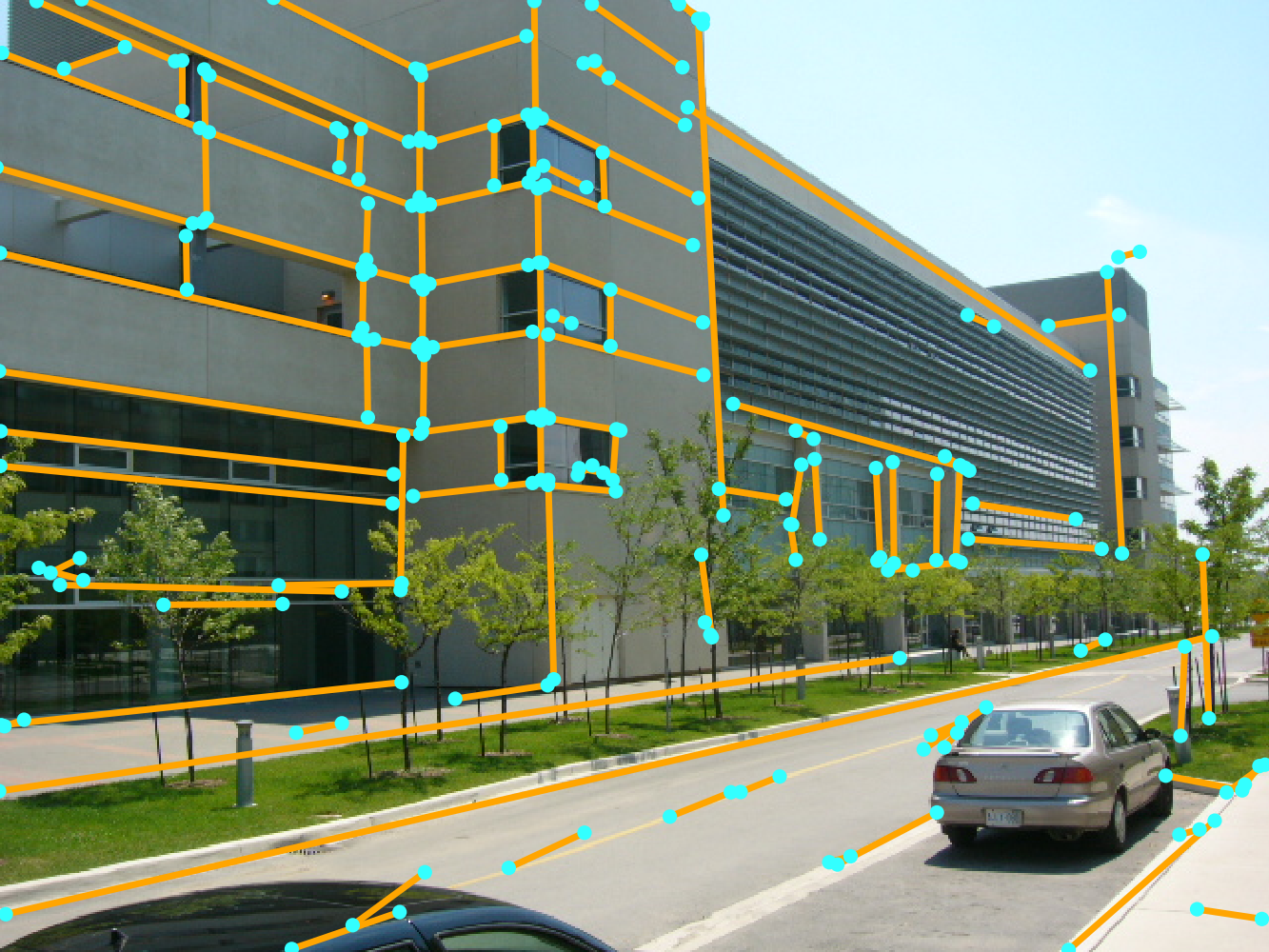}
     \end{subfigure}
     \begin{subfigure}[b]{0.15\textwidth}
         \centering
         \includegraphics[width=1\linewidth]{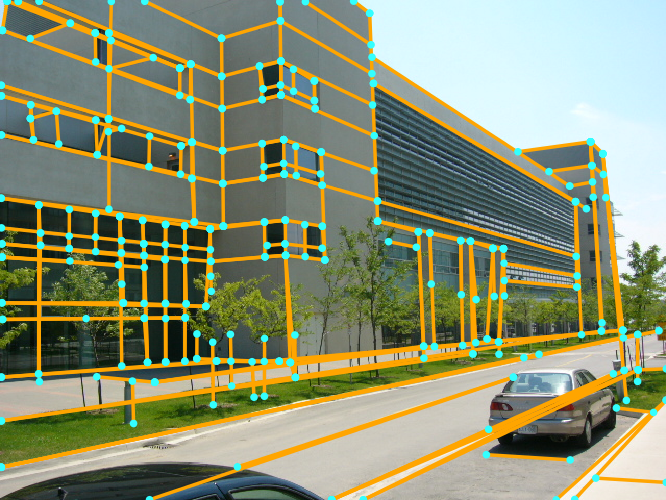}
     \end{subfigure}
     \begin{subfigure}[b]{0.15\textwidth}
         \centering
         \includegraphics[width=1\linewidth]{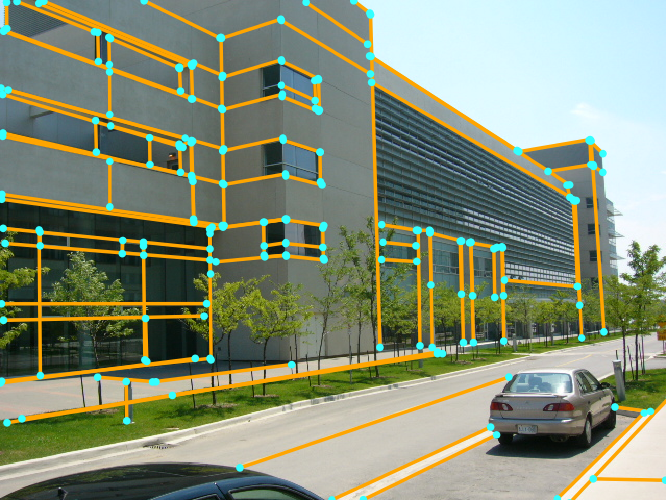}
     \end{subfigure}
     \begin{subfigure}[b]{0.15\textwidth}
         \centering
         \includegraphics[width=1\linewidth]{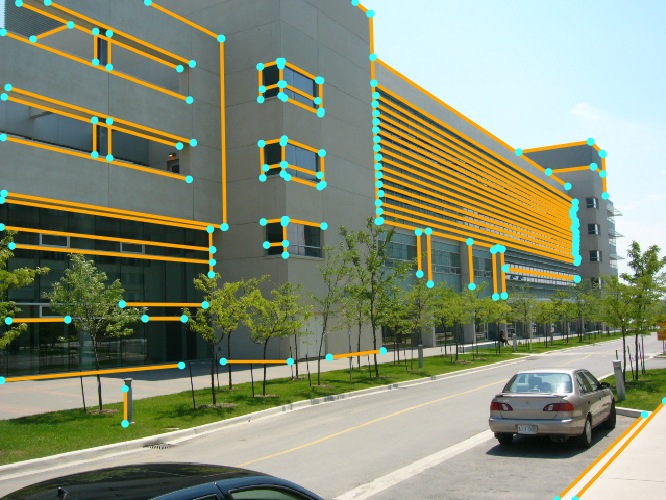}
     \end{subfigure}
     ~~
     
     \begin{subfigure}[b]{0.15\textwidth}
         \centering
         \includegraphics[width=1\linewidth]{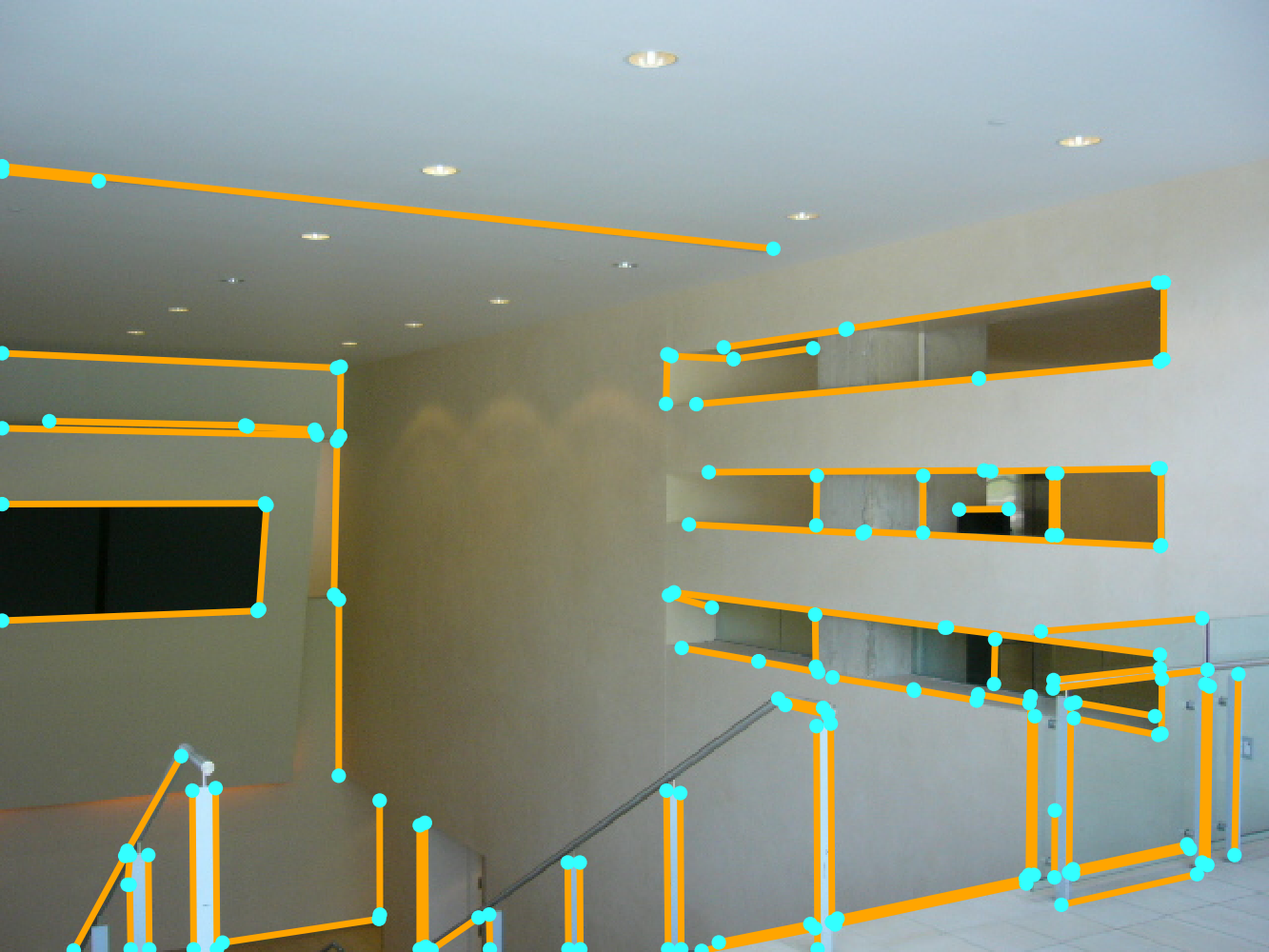}
     \end{subfigure}
    \begin{subfigure}[b]{0.15\textwidth}
         \centering
         \includegraphics[width=1\linewidth]{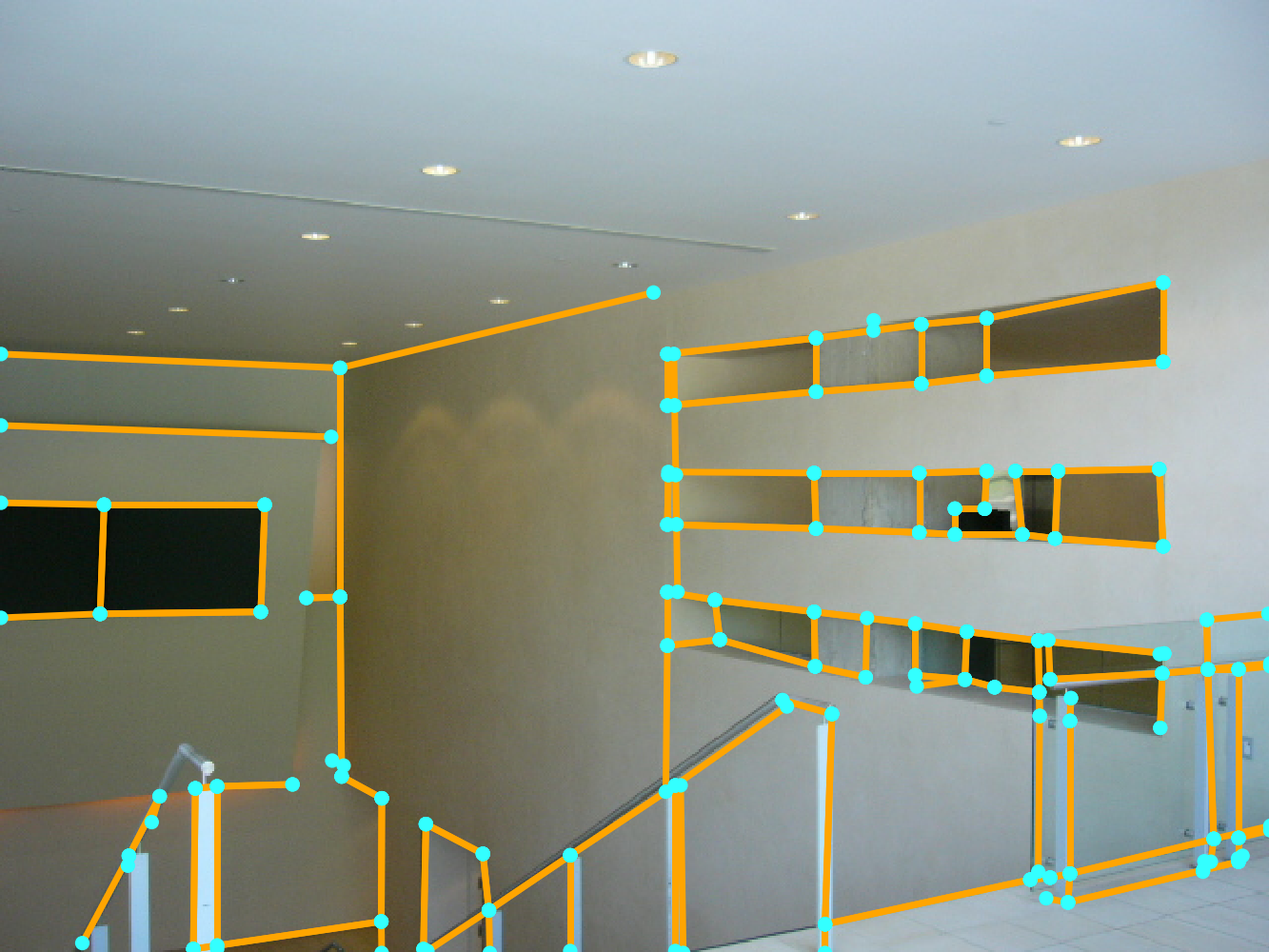}
     \end{subfigure}
     \begin{subfigure}[b]{0.15\textwidth}
         \centering
         \includegraphics[width=1\linewidth]{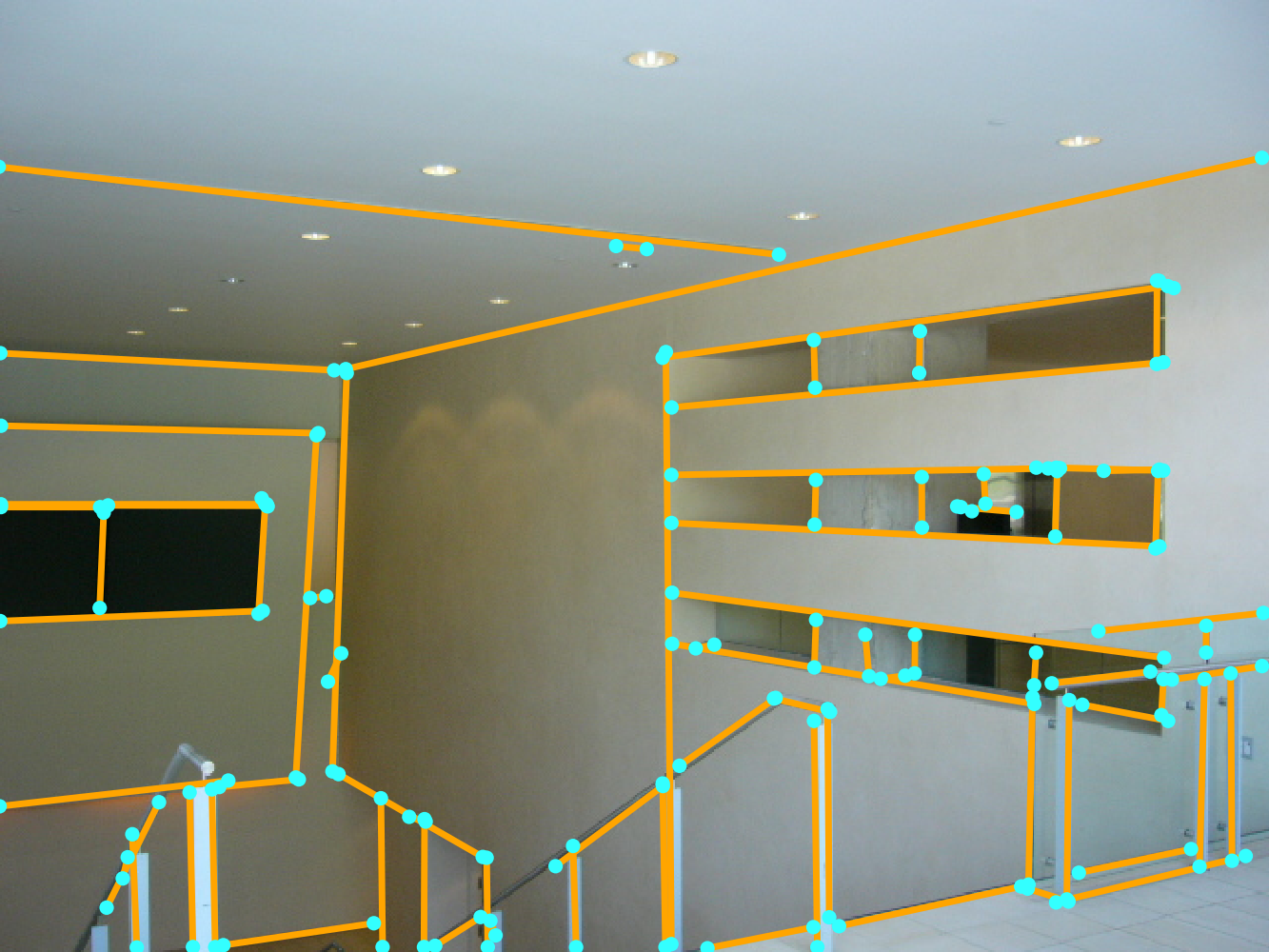}
     \end{subfigure}
     \begin{subfigure}[b]{0.15\textwidth}
         \centering
         \includegraphics[width=1\linewidth]{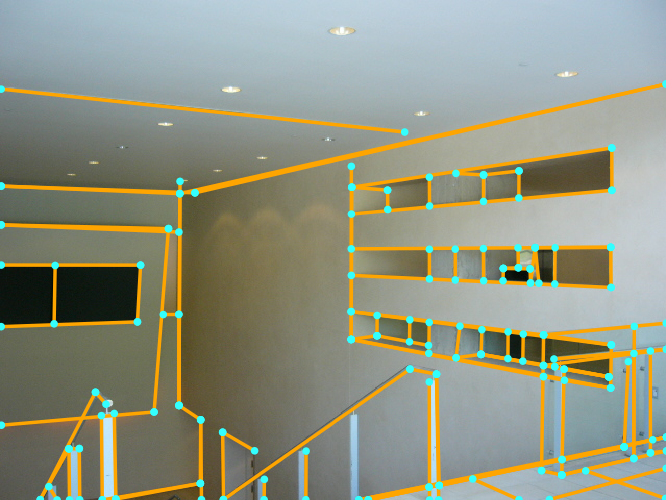}
     \end{subfigure}
     \begin{subfigure}[b]{0.15\textwidth}
         \centering
         \includegraphics[width=1\linewidth]{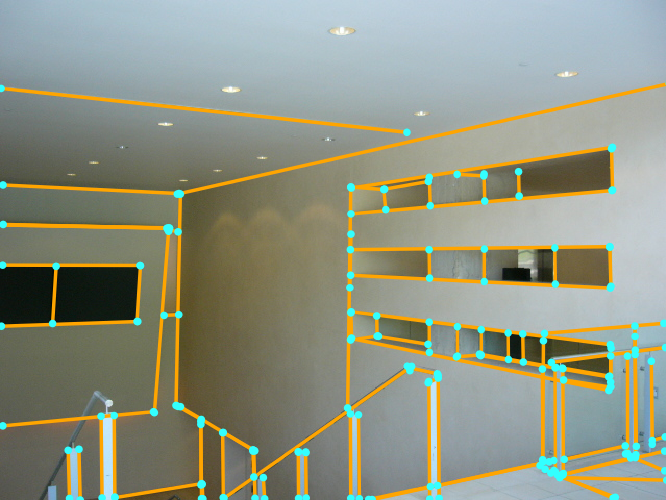}
     \end{subfigure}
     \begin{subfigure}[b]{0.15\textwidth}
         \centering
         \includegraphics[width=1\linewidth]{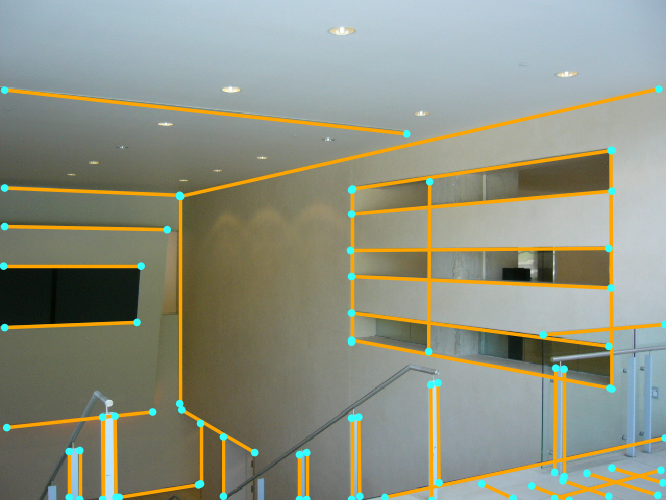}
     \end{subfigure}
     ~~
     
      \begin{subfigure}[b]{0.15\textwidth}
         \centering
         \includegraphics[width=1\linewidth]{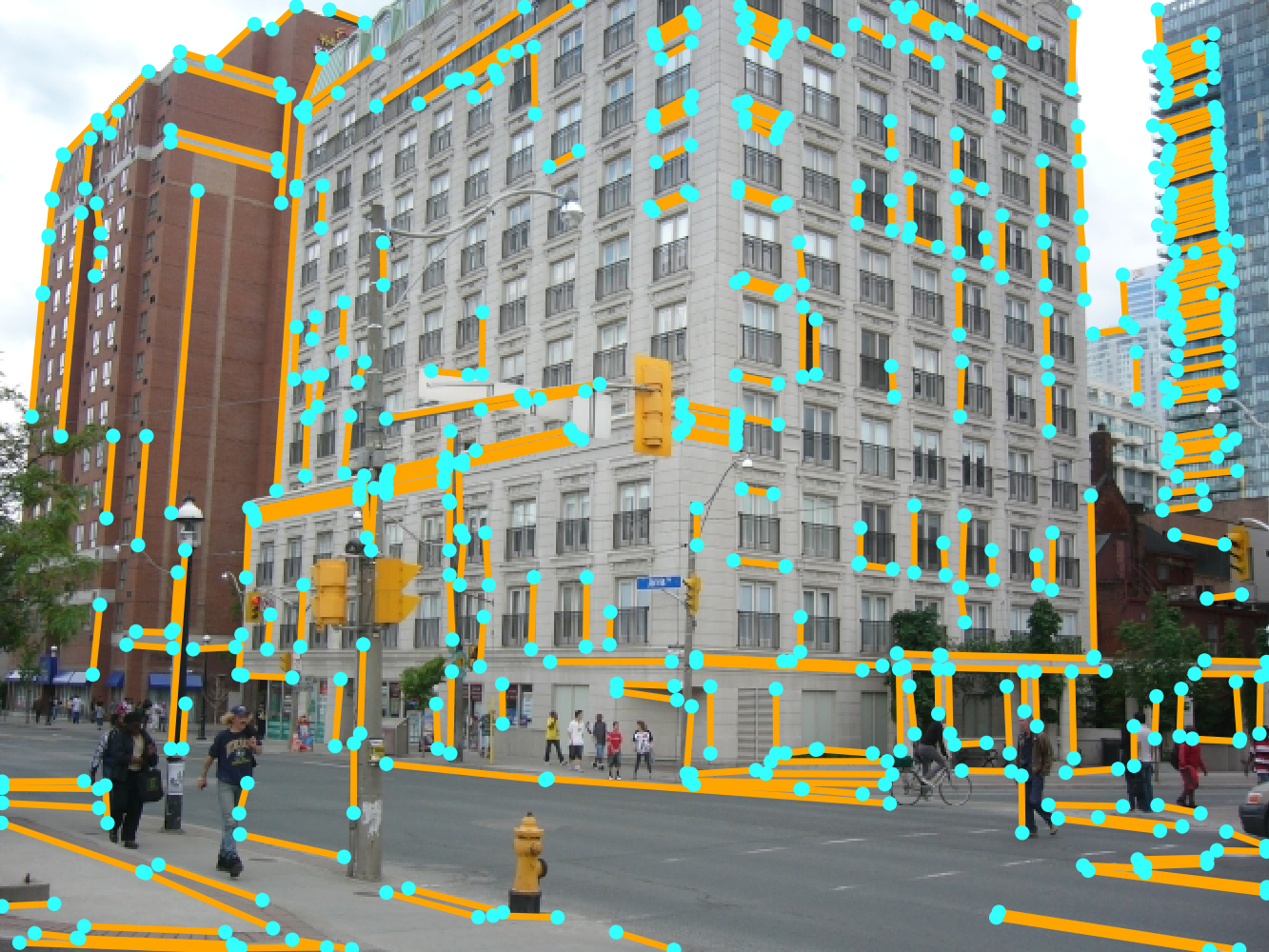}
         \caption{LSD}
     \end{subfigure}
    \begin{subfigure}[b]{0.15\textwidth}
         \centering
         \includegraphics[width=1\linewidth]{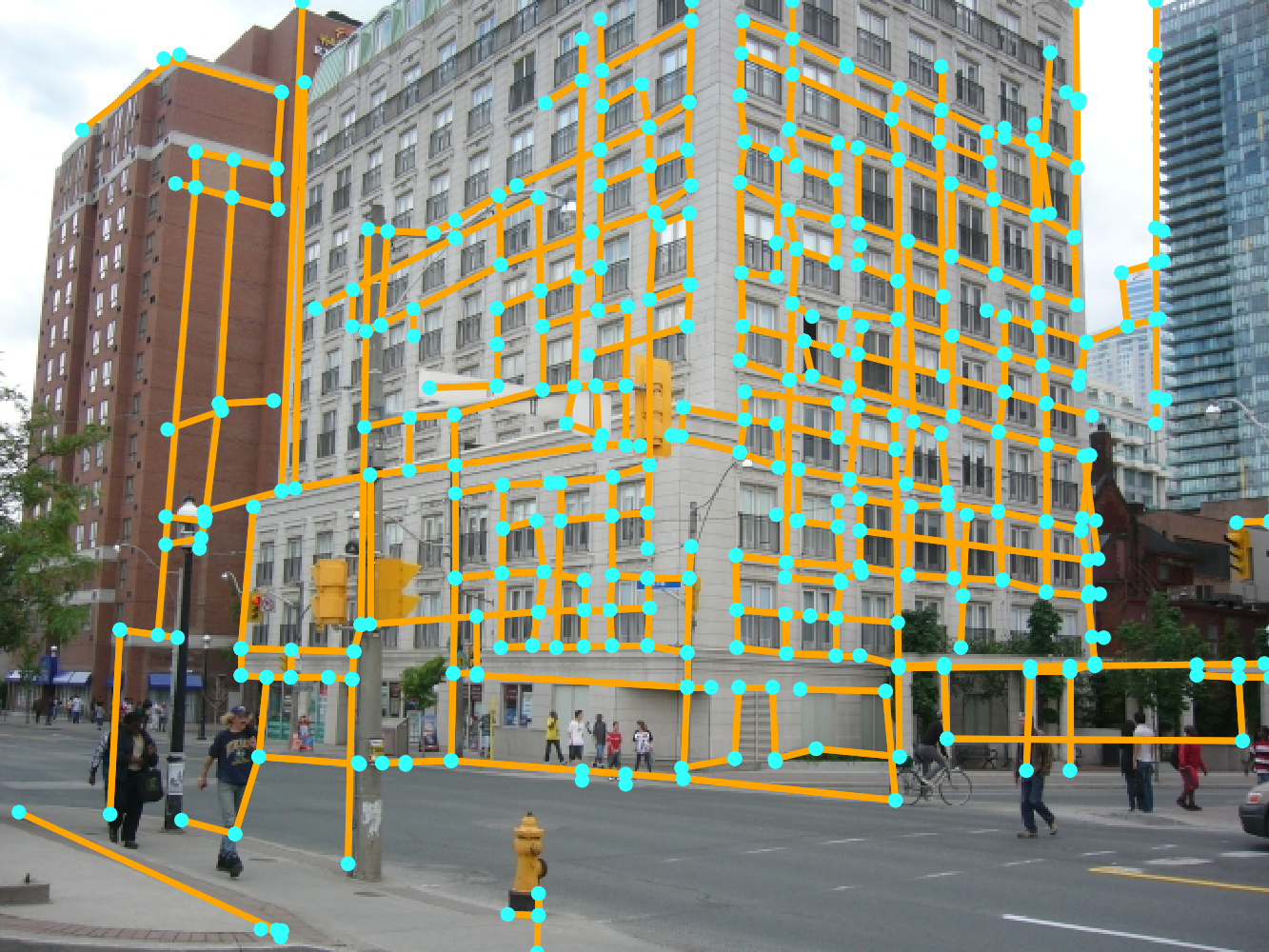}
         \caption{DWP}
     \end{subfigure}
     \begin{subfigure}[b]{0.15\textwidth}
         \centering
         \includegraphics[width=1\linewidth]{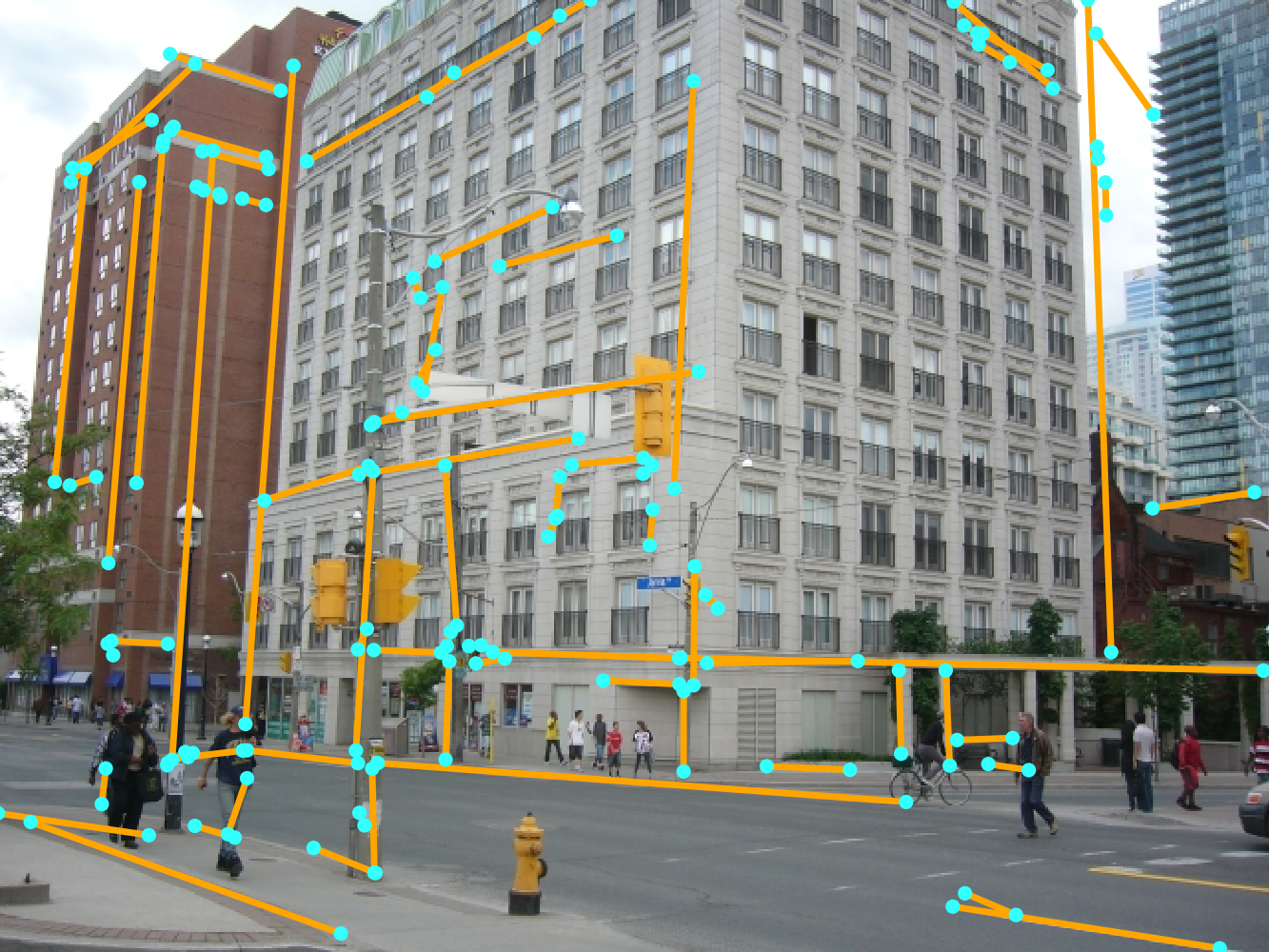}
         \caption{AFM}
     \end{subfigure}
     \begin{subfigure}[b]{0.15\textwidth}
         \centering
         \includegraphics[width=1\linewidth]{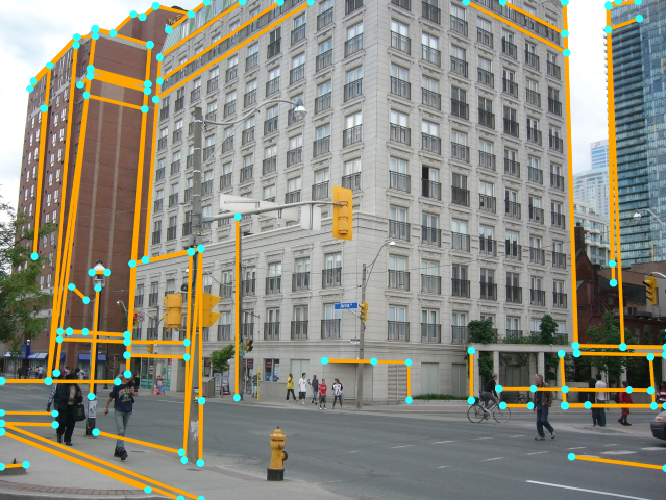}
         \caption{L-CNN}
     \end{subfigure}
     \begin{subfigure}[b]{0.15\textwidth}
         \centering
         \includegraphics[width=1\linewidth]{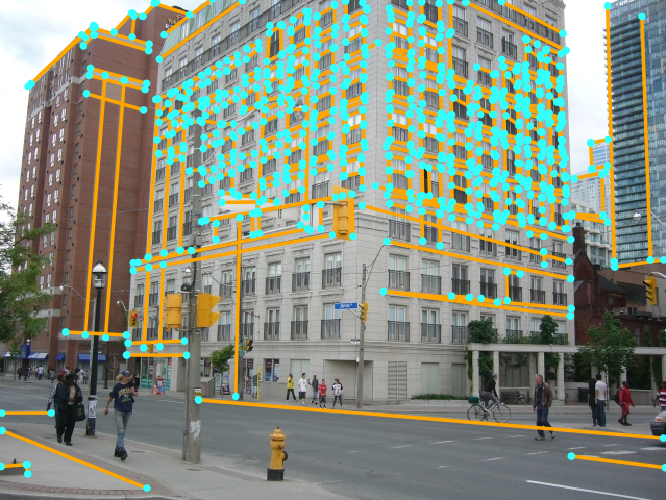}
         \caption{Ours}
     \end{subfigure}
     \begin{subfigure}[b]{0.15\textwidth}
         \centering
         \includegraphics[width=1\linewidth]{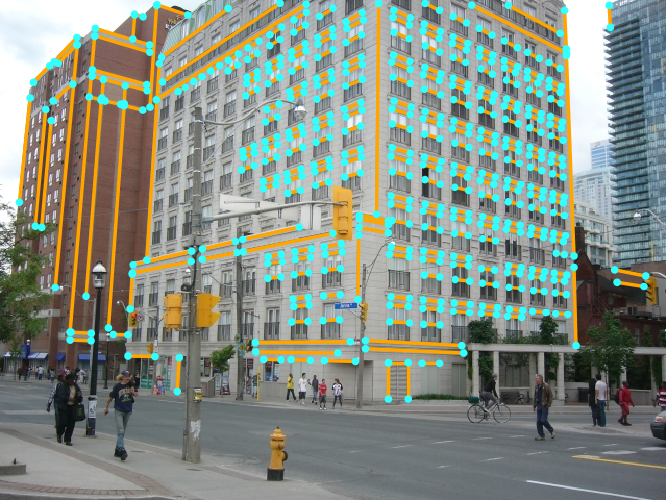}
         \caption{GT}
     \end{subfigure}
     ~~
     
   \end{center}
    \caption{Quanlitative evaluation of line detection methods on Wireframe dataset and YorkUrban dataset. The line segments and their end-points are marked by orange and cyan colors, respectively.}
    \label{vis_pic}
\end{figure}

\textbf{Visualization and Discussion}
In Fig.~\ref{vis_pic}, several results of line segment detection are visualized.
LSD detected some noisy local textures without semantic meaning. Recent CNN-based methods have shown good noise-suppression ability because they obtain high-level semantics. AFM does not have explicit endpoint definition, limiting the accuracy of end-points localization. It also presented many short line segments.
DWP gives a relatively cleaner detection result, but there exist some incorrectly connected junction pairs, caused by inaccurate junctions predictions and sub-optimal heuristic combination algorithm. L-CNN, which has a junction detector and an extra line segment classifier, has good visualization results. However, its line segment detection result rely on the junction detection and line feature sampling, which might be prone to missed junction and nearby texture variation. In comparison, the proposed TP-LSD method is capable to detect line segments in complicated even low-contrast environments as is shown on the 
first and the sixth rows in Fig.~\ref{vis_pic}.

\textbf{Inference Speed}
Based on NVIDIA RTX2080Ti GPU and Intel Xeon Gold 6130 2.10 GHz CPU, the inference speed is reported in Table \ref{eval_table}. With the image size of $320\times 320$, the proposed TP-LSD achieve the real-time speed up to 78 FPS, offering the potential to be used in real-time applications like SLAM.

\section{Conclusion}
This paper proposes a faster and more compact model TP-LSD for line segment detection with the one-step strategy. Tri-points representation is used to encodes a line segment with three keypoints, based on which the line-segment detection is realized by end-to-end inference.
Furthermore, the straight line-map is produced based on segmentation task, and is used as structural prior cues to guide the extraction of TPs.
Both quantitatively and qualitatively, TP-LSD shows the improved performances compared to the existing models. Besides, 
our method achieves 78 FPS speed, showing potential to be integrated with real-time applications, such as vanishing point estimation, 3D reconstruction and pose estimation.

\newpage
\bibliographystyle{splncs04}
\bibliography{main}

\end{document}